\documentclass[11pt]{article}
\usepackage[dvipsnames]{xcolor}

\usepackage[framemethod=default]{mdframed}
\usepackage{caption}

\usepackage{indentfirst}
\usepackage{bm, mathrsfs, graphics,float,amssymb,amsmath,subeqnarray,setspace,graphicx,amsthm,epstopdf,subfigure, enumerate, color}
\usepackage[utf8]{inputenc}
\usepackage[colorlinks,
            linkcolor=red,
            anchorcolor=blue,
            citecolor=blue
            ]{hyperref}
\usepackage{natbib}

\usepackage{fullpage}
\usepackage{multirow}
\usepackage{tablefootnote}
\usepackage{colortbl}% http://ctan.org/pkg/xcolor
\usepackage{hhline}% http://ctan.org/pkg/hhline
\usepackage[titletoc]{appendix}

\usepackage{algorithm}
\usepackage{algorithmic}

\usepackage{mathtools}

\begin{document}
\title{
ADA-Tucker: Compressing Deep Neural Networks via Adaptive Dimension Adjustment Tucker Decomposition
}

\author{
Zhisheng Zhong
\thanks{
Peking University;
email: \{zszhong, weifangyin, zlin\}@pku.edu.cn, chzhang@cis.pku.edu.cn
}
\qquad
Fangyin Wei
\footnotemark[1]
\qquad
Zhouchen Lin
\footnotemark[1]
\qquad
Chao Zhang
\footnotemark[1]
\thanks{
	Corresponding author
}
}
\date{}

\maketitle
%\tableofcontents

\begin{abstract}
Despite recent success of deep learning models in numerous applications, their widespread use on mobile devices is seriously impeded  by storage and computational requirements. In this paper, we propose a novel network compression method called Adaptive Dimension Adjustment Tucker decomposition (ADA-Tucker). With learnable core tensors and transformation matrices, ADA-Tucker performs Tucker decomposition of \textit{arbitrary-order} tensors. Furthermore, we propose that weight tensors in networks with proper order and balanced dimension are easier to be compressed. Therefore, the high flexibility in decomposition choice distinguishes ADA-Tucker from all previous low-rank models. To compress more, we further extend the model to Shared Core ADA-Tucker (SCADA-Tucker) by defining a shared core tensor for all layers. Our methods require no overhead of recording indices of non-zero elements. Without loss of accuracy, our methods reduce the storage of LeNet-5 and LeNet-300 by ratios of $\mathbf{691\times}$ and $\mathbf{233\times}$, respectively, significantly outperforming state of the art. The effectiveness of our methods is also evaluated on other three benchmarks (CIFAR-10, SVHN, ILSVRC12) and modern newly deep networks (ResNet, Wide-ResNet).
\end{abstract}

\section{Introduction}

Driven by increasing computation power of GPUs and huge amount of data, deep learning has recently made great achievements in computer vision, natural language processing and speech recognition. In the history of neural network \cite{lecun1998gradient,krizhevsky2012imagenet,simonyan2014very,szegedy2015going,he2016deep,huang2016densely}, networks tend to have more layers and more weights. Although deeper neural networks may achieve better results, the expense of storage and computation is still a great challenge. Due to limits of devices and increasing demands from many applications, effective network compression for convolutional (Conv) layers and fully-connected (FC) layers is a critical research topic in deep learning. 

\begin{figure}%{tbhp} 
	\centering
	\includegraphics[width=1\linewidth]{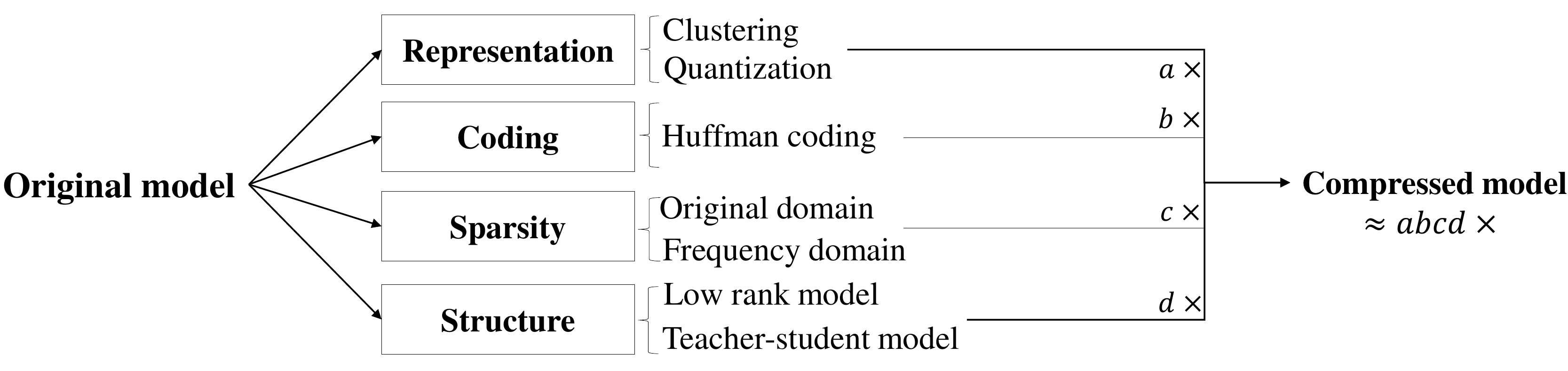}
	\caption{Four categories of mainstream compression methods, which can be combined for higher compression ratio. ``$a\times$'' etc. means that the network is compressed by $a$ times.}
	\label{flowchart}   
\end{figure}

So far, as illustrated in Figure \ref{flowchart}, mainstream methods for network compression can be categorized into four groups: reducing the bits of weight representation, effective coding, making weights sparse and simplifying the network structure. These four methods can be combined together for higher compression ratio with little loss in network performance. Han et al. have combined the first three methods in \cite{han2015deep}.

\paragraph{Reducing the bits of weight representation and Effective coding} There are two approaches for the first category: clustering and quantization. BinaryConnect \cite{courbariaux2015binaryconnect} enforces weights in neural networks to take binary values. Incremental network quantization \cite{zhou2017incremental} quantizes  deep models with 5 bits incrementally. Gong et al. \cite{gong2014compressing} learn CNNs in advance, and then apply k-means clustering on the weights for quantization.  Ullrich et al. \cite{ullrich2017soft} cluster the weights with a Gaussian mixture model (GMM), using only six class centers to represent all weights. The second category, effective coding, always combines with the first category, where the coding scheme is mainly Huffman coding. DeepCompression \cite{han2015deep} first introduces Huffman coding in network compression and improves the compression ratios further. CNNPack \cite{wang2016cnnpack} also uses Huffman coding and gets better results.

\paragraph{Making weights sparse} Sparsity can be induced in either the original domain or the frequency domain. The most commonly used sparsity method in the original domain is pruning. Han et al. \cite{han2015learning} recursively train a neural network and prune unimportant connections based on their weight magnitude. Dynamic network surgery \cite{guo2016dynamic} prunes and splices the branch of the network. The frequency domain sparsity methods benefit from discrete cosine transformation (DCT). Chen et al. \cite{chen2016compressing} take advantage of DCT to make weights sparse in the frequency domain. Wang et al. \cite{wang2016cnnpack} combine DCT, clustering and Huffman coding for further compression.

\begin{figure*}
	\centering
	\includegraphics[width=1\textwidth]{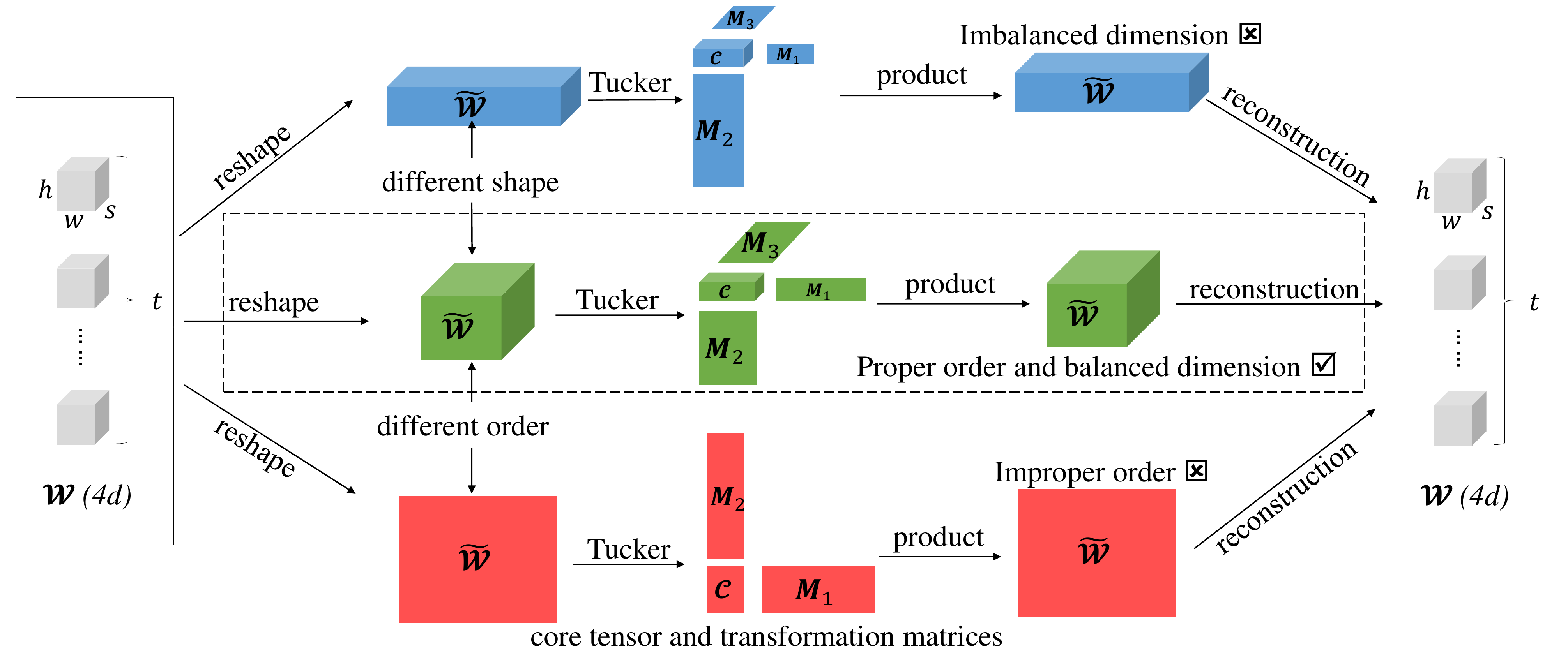}
	\caption{Illustration of ADA-Tucker: For each layer, the order for Tucker decomposition depends on the dimensions of modes of the original tensor. For different layers, orders and dimensions of tensors can vary.}
	\label{tucker_illustration_ada}
\end{figure*}

\begin{figure*}
	\centering
	\includegraphics[width=1\textwidth]{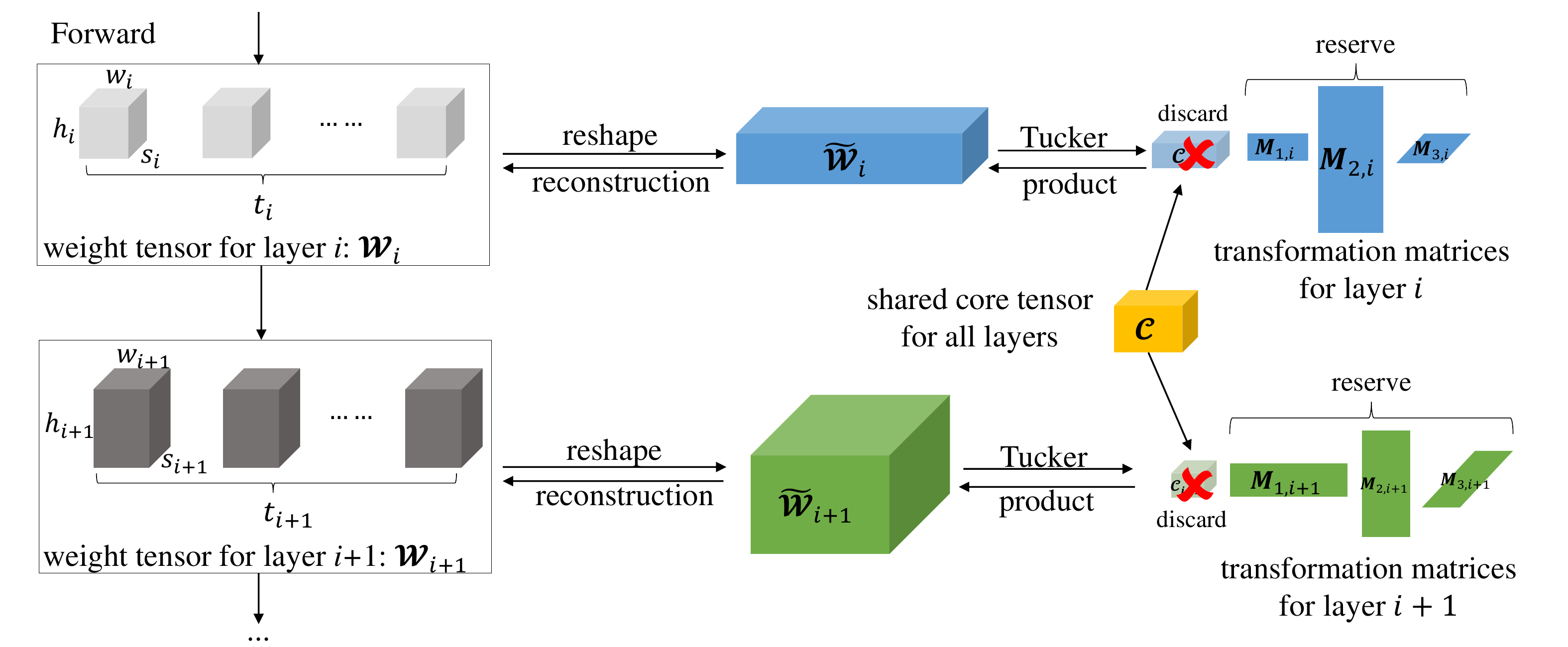}
	\caption{Illustration of SCADA-Tucker: If all layers share the same core tensor, i.e., $\forall i\in{1,2,...,l},  \boldsymbol{\mathcal{C}}^{(i)}=\boldsymbol{\mathcal{C}}$,  ADA-Tucker becomes SCADA-Tucker.}
	\label{tucker_illustration_scada}
\end{figure*}

\paragraph{Simplifying the network structure} A common approach of the fourth category involves matrix and tensor decomposition, while another rarely used approach called teacher-student model \cite{ba2014deep,hinton2015distilling} tries to reduce the depth of networks. Low-rank models were first used in the fully-connected layer \cite{denil2013predicting}. They utilize singular value decomposition (SVD) to reduce the computation and storage. Tensor Train decomposition \cite{novikov2015tensorizing} is another model to compress fully-connected layer. Jaderberg et al. \cite{jaderberg2014speeding}, Denton et al. \cite{denton2014exploiting} and Tai et al. \cite{tai2015convolutional} speed up CNNs with low-rank regularization. Canonical Polyadic \cite{lebedev2014speeding} and Tucker decomposition \cite{kim2015compression} are advocated to accelerate the training of CNNs.

Our model falls into the last category, and it differs from the existing methods in two folds. First, while previous methods generally decompose weight tensors with fixed order and dimension, our methods adaptively adjust the original weight tensor into a new tensor with arbitrary order before Tucker decomposition. The superiority of such flexibility will be explained and demonstrated in the next section. Second, the proposed model can be applied to \textit{both Conv and FC layers}, requiring no definition of new layers. In fact, previous low-rank models implemented by defining additional layers are special cases of our methods.

In principle, our methods can also combine with other three categories for higher compression ratios. In the experiments section, we combine quantization and Huffman coding for better results. 

In summary, our contributions are as follows:

\begin{itemize}
	\item We demonstrate that deep neural networks can be better compressed using weight tensors with proper orders and balanced dimensions of modes without performance degradation.
	
	\item We propose a novel network compression method called ADA-Tucker with flexible decomposition that drastically compresses deep networks while learning.
	
	\item We further extend ADA-Tucker to SCADA-Tucker with a shared core tensor for all layers, achieving even higher compression ratios with negligible accuracy loss.
	
\end{itemize}

\section{ADA-Tucker and SCADA-Tucker}\label{ada}

\textbf{Notations:} Following \cite{kolda2009tensor}, tensors are denoted by boldface Euler script letters, e.g., $\boldsymbol{\mathcal{A}}$, matrices are denoted by boldface capital letters, e.g., $\boldsymbol{A}$, vectors are denoted by boldface lowercase letters, e.g., $\boldsymbol{a}$, and scalars are denoted by lowercase letters, e.g., $a$. $\boldsymbol{\mathcal{A}}^{(i)}$ represents the parameters of the $i$-th layer and $\boldsymbol{A}_{(i)}$ represents the $i$-mode of tensor $\boldsymbol{\mathcal{A}}$.

% Vectors are denoted by boldface lowercase letters, e.g., $\textbf{a}$.

\subsection{Tensor Decomposition on the Weight Tensor}\label{ADA-Tucker conv}

Weights of a deep neural network mainly come from Conv layers and FC layers. With weights in both types of layer represented by tensors, methods based on tensor decomposition can be applied to reduce the weight numbers.

For a Conv layer, its weight can be represented by a fourth order tensor $\boldsymbol{\mathcal{W}}\in \mathbb{R}^{h \times w \times s \times t}$, where $h$ and $w$ represent the height and width of the kernel, respectively, and $s$ and $t$ represent the channel number of input and output, respectively. Similarly, the weight of a FC layer can be viewed as a second order tensor $\boldsymbol{\mathcal{W}}\in \mathbb{R}^{s \times t}$, where $s$ and $t$ represent the number of the layer's input and output units, respectively. Thus in general, the form of a weight tensor is a $d_w$-th order $(m_1,m_2,...,m_{d_w})$-dimensional tensor  $\boldsymbol{\mathcal{W}}\in \mathbb{R}^{m_1\times m_2\times ...  \times m_{d_w}}$, where $m_i$ is the dimension of the $i$-th mode.

The weight tensor can be original if the magnitude of $m_i$'s is balanced. Otherwise, it can be a reshaped version of the original tensor according to the adaptive dimension adjustment mechanism described in the next subsection. Suppose that $\boldsymbol{\mathcal{W}}$ is reshaped into $\boldsymbol{\mathcal{\tilde{W}}}\in \mathbb{R}^{n_1\times n_2 \times ... \times n_{d_c}}$, where $n_1\times n_2\times ... \times n_{d_c}=m_1\times m_2\times ... \times m_{d_w}$. Then based on Tucker decomposition, we decompose the reshaped weight tensor $\boldsymbol{\mathcal{\tilde{W}}}$ into a $d_c$-mode product of a core tensor $\boldsymbol{\mathcal{C}}$ and a series of transformation matrices $\{\boldsymbol{M}\}$: 
\begin{equation}\label{tucker_any}
\boldsymbol{\mathcal{\tilde{W}}}\approx \boldsymbol{\mathcal{C}}\times_1 \boldsymbol{M}_1 \times_2 \boldsymbol{M}_2 \times_3 ... \times_{d_c} \boldsymbol{M}_{d_c}
,\end{equation}
where $\boldsymbol{\mathcal{C}}\in \mathbb{R}^{k_1\times k_2\times ...  \times k_{d_c}}$ and $\boldsymbol{M}_i \in \mathbb{R}^{n_i \times k_i}         (i=1,2,...,d_c)$ are all learnable. They need to be stored during training in order to reconstruct $\boldsymbol{\mathcal{W}}$: after the $d_c$-mode product w.r.t. $\boldsymbol{\mathcal{C}}$, we reshape ${\boldsymbol{\mathcal{\tilde{W}}}}$ into  $\boldsymbol{\mathcal{W}}$ so as to produce the output of the layer in forward propagation and pass the gradients in backward propagation.

We define $\boldsymbol{\tilde{W}}_{(i)}\in \mathbb{R}^{n_i\times (n_1... n_{i-1}n_{i+1}... n_{d_c})}$ and $\boldsymbol{C}_{(i)}\in \mathbb{R}^{k_i\times (k_1... k_{i-1}k_{i+1}... k_{d_c})}$ as the $i$-mode unfolding of tensor $\boldsymbol{\mathcal{\tilde{W}}}$ and $\boldsymbol{\mathcal{C}}$, respectively, and rewrite Eq.~\eqref{tucker_any} as:
\begin{eqnarray}\label{mode_n_tucker}
\begin{aligned}
\boldsymbol{\tilde{W}}_{(i)} = 
\boldsymbol{M}_i\boldsymbol{C}_{(i)}\left(\boldsymbol{M}_{d_c}\otimes\boldsymbol{M}_{d_c-1}\otimes...\otimes\boldsymbol{M}_{i+1}\otimes\boldsymbol{M}_{i-1}\otimes...\otimes\boldsymbol{M}_{1}\right)^T,
\end{aligned}\end{eqnarray}
where $\otimes$ represents the Kronecker product. The gradients of loss $L$ w.r.t. the core tensors and the transformation matrices are as follows:

\begin{eqnarray}\label{dm}
\begin{aligned}
\frac{\partial L}{\partial \boldsymbol{M}_i} =
\frac{\partial L}{\partial \boldsymbol{\tilde{W}}_{(i)}}
\left(\boldsymbol{M}_{d_c}\otimes\boldsymbol{M}_{d_c-1}\otimes...\otimes\boldsymbol{M}_{i+1}\otimes\boldsymbol{M}_{i-1}\otimes...\otimes\boldsymbol{M}_{1}\right)\boldsymbol{C}_{(i)}^T
,
\end{aligned}\end{eqnarray}

\begin{eqnarray}\label{dc}
\begin{aligned}
\frac{\partial L}{\partial \boldsymbol{C}_{(i)}} =
\boldsymbol{M}_i^T\frac{\partial L}{\partial \boldsymbol{\tilde{W}}_{(i)}}\left(\boldsymbol{M}_{d_c}\otimes\boldsymbol{M}_{d_c-1}\otimes...\otimes \boldsymbol{M}_{i+1}\otimes\boldsymbol{M}_{i-1}\otimes...\otimes\boldsymbol{M}_{1}\right),
\end{aligned}
\end{eqnarray}

\begin{eqnarray}\label{dc_reshape}
\frac{\partial L}{\partial \boldsymbol{\mathcal{C}}} =
{\rm{fold}}\left(\frac{\partial L}{\partial \boldsymbol{C}_{(i)}}\right).
\end{eqnarray}

\subsection{Adaptive Dimension Adjustment and motivation}\label{dimension_adjust}

The tendency of network overfitting suggests that there is always redundancy among the weights which can be approximately measured by `rank'. And `rank' is often determined by the smaller or smallest size of different modes (e.g., for a matrix, its rank cannot exceed its row number or column number, whichever smaller). If the size of a mode is much smaller than others, compressing along that mode will cause significant information loss. 

Changing the dimension of the weight tensors to avoid the significant information loss in DNNs has been widely used in network compression. For example, Jaderberg et al. \cite{jaderberg2014speeding} compress network with low-rank regularization. In their model, they merged the kernel height dimension and kernel width dimension into one dimension and got success, which suggests that there exist some information redundancy between the kernel width dimension and kernel height dimension. ThiNet \cite{luo2017thinet} proposed to compress the weights through the input channel dimension and output channel dimension, which suggests that there exist some information redundancy between the input channel dimension and output channel dimension. Zhang et al. \cite{zhang2017interleaved} proposed interleaved group convolutions that splitting the weight tensor into several small group tensor, which also means that there exist redundancy among the four dimensions. Here, we extend these ideas further. We treat all four dimensions of the weight tensor equally. So we reshape the weight tensor to any order and any shape. Here is a toy example that can illustrate this idea. Suppose that we have 100 parameters represented by a matrix of size $1\times100$ or $10\times10$. Obviously, the rank of the former matrix tends to be 1, in which case rank-based compression is hard (compressing to a zero-rank matrix will lose all information). In contrast, a matrix of real data in the latter form can be easily approximated with a lower-rank matrix without losing too much information.

As a conclusion, compression will be much less effective if a tensor is not reshaped to one with appropriate order and balanced dimension. Motivated by such consideration, we implement adaptive dimension adjustment in our model that allows reshaping weight tensors and defining core tensors of arbitrary shape. Experiments also demonstrate that both balanced dimensions of each mode and a proper order of the weight tensor contribute to better performance during network compression. 

In the following subsection, we will describe the principle and process of adaptive dimension adjustment.

\subsubsection{Adaptive Dimension Adjustment for Conv Layers}\label{ada_conv} 

For a Conv layer, the basic mechanism is to reshape the original weight tensor into a tensor with roughly even dimensions of modes. We take the Conv1 (first convolutional) layer of LeNet-5 as an example. The size of its original weight tensor is $5\times5\times1\times20$. Normally, a mode of dimension one is redundant and can be simply neglected (such case usually occurs in the first convolutional layer of a neural network). Note that dimensions of the first two modes are much smaller than that of the last one. With $20$ still an acceptable dimension size, we merge the first two modes and get a second order tensor of size $25\times20$. We may then define a smaller second order core tensor accordingly for decomposition. 

Generally speaking, when there are few input channels (e.g., for the first layer of a network, $s=1$ or $s=3$), we merge the input and output channels into one mode, obtaining a third order weight tensor $\boldsymbol{\mathcal{\tilde{W}}}\in \mathbb{R}^{h \times w \times st}$. Similar operation is conducted for small kernel size (e.g., $1 \times 1$ or $5 \times 5$), i.e., merging the first two modes into one to have $\boldsymbol{\mathcal{\tilde{W}}}\in \mathbb{R}^{hw \times s\times t}$. If these two cases occur simultaneously, we can reduce the original fourth order weight tensor to a matrix $\boldsymbol{\mathcal{\tilde{W}}}\in \mathbb{R}^{hw \times st}$. With the same principle in mind, when the kernel size grows large enough, it is better to maintain the weight tensor as a fourth order tensor or even reshape it to one with a higher order (e.g., fifth order and sixth order). In fact, the dimension adjustment operations are \textit{not limited to simply merging several modes}: any form of reshaping operation is valid as long as the number of weight stays the same, which, as far as we know, is not achieved by any previous low-rank models.

\begin{figure*}[tb]
	\centering
	\includegraphics[width=0.85\linewidth]{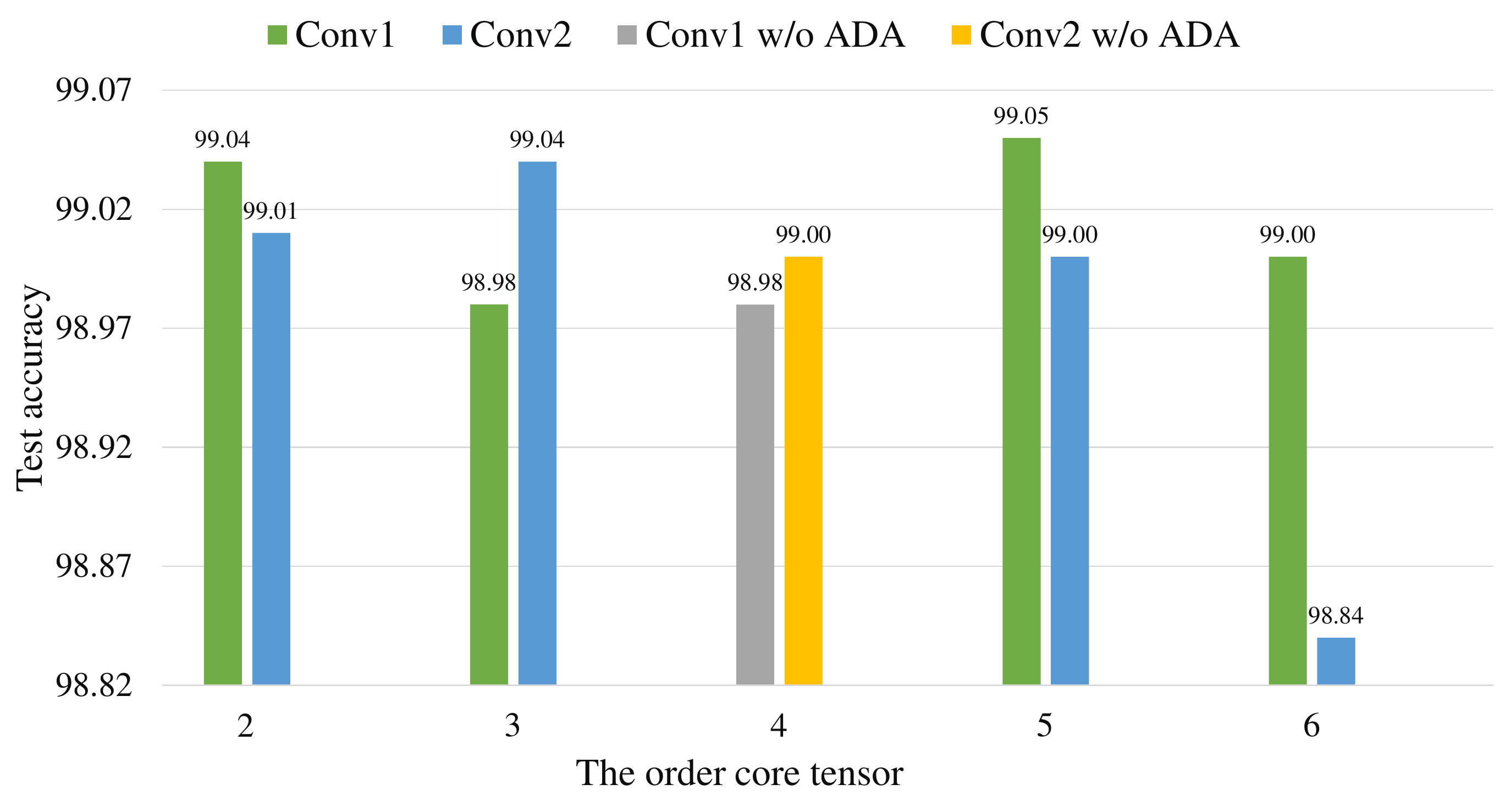}
	\caption{Dimension adjustment of Conv layers. Bars of the same color represent experiments with similar numbers of weight, conducted by changing the order of the weight tensor of a specific layer while fixing the rest. The optimal order for Conv1 and Conv2 layers in LeNet-5 are five and three, respectively (better viewed together with Table \ref{conv1} and Table \ref{conv2}). }
	\label{dimension_conv}  
\end{figure*}

We designed experiments about adaptive dimension adjustment of Conv1 and Conv2 layers in LeNet5. We conducted this experiments by changing the order of the weight tensor of Conv1/Conv2 layer while fixing the rest. The details of the Conv1/Conv2’s weight tensor with different orders are listed in Table  \ref{conv1} and Table \ref{conv2} of the appendices part. We chose proper core tensor sizes for each order to ensure the numbers of parameters under different settings are similar. The network performances under different settings are showing in Figure \ref{dimension_conv}. From Figure \ref{dimension_conv}, the optimal order for Conv1 and Conv2 layers in LeNet-5 are five and three, respectively. Here is one more important thing to mention, the gray and yellow bars mean we did not use adaptive dimension adjustment on these two settings. The original order for Conv layer's weight is four, so our ADA-Tucker degenerates to Tucker under these two settings. From the results, if we reshape the tensor with proper order and balanced dimensions before Tucker decomposition, we can get better compressed results.

\subsubsection{Adaptive Dimension Adjustment for FC Layers}\label{ada_fc} 
Our dimension adjustment mechanism also applies to FC layers. Here we present an example involving a fifth order reshaped tensor for FC2 (second fully-connected) layer of LeNet-5, which is originally a matrix of size $500 \times 10$. To balance the dimensions of modes, we reshape the original weight tensor to a size of $5 \times 5 \times 5 \times 5 \times 8$. Note that such operation does not necessarily indicate splitting individual mode, the decomposition is allowed to disrupt the original sequence of dimension size (e.g., 8 is a factor of neither 500 nor 10). With the weight tensor reshaped as a fifth order tensor according to our adaptive dimension adjustment principle, the network finds its best structure for the FC2 layer. To our knowledge, previous tensor decomposition methods can only regard FC layer's weights as a second order tensor with fixed dimensions of modes.

\begin{figure*}[tb]
	\centering
	\includegraphics[width=0.85\linewidth]{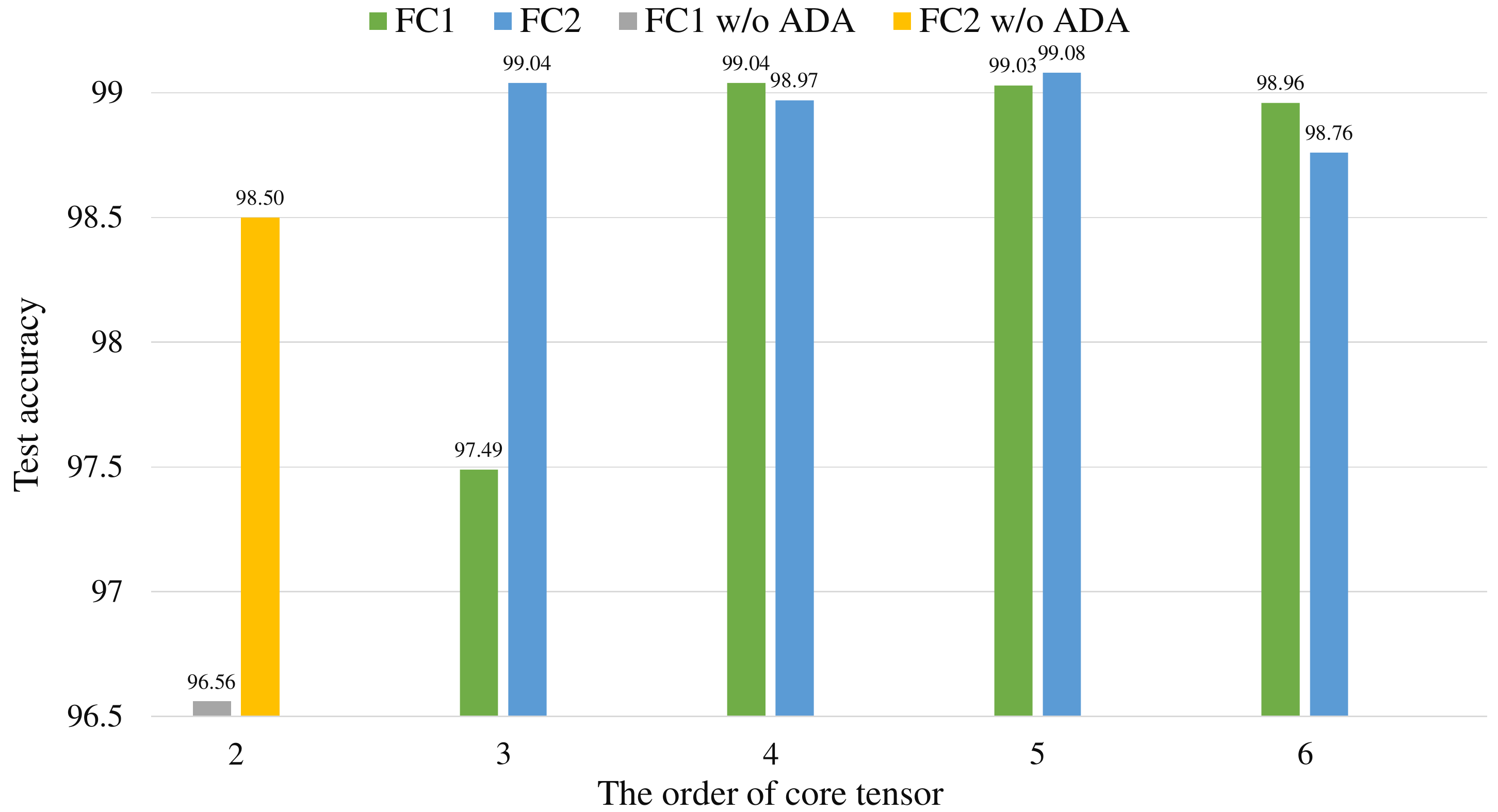}
	\caption{Dimension adjustment of FC layers. Bars with same color represent experiments with similar numbers of weight, conducted by changing the order of the weight tensor of a specific layer while fixing the rest. A fourth order weight tensor is optimal for FC1 layer while a fifth order weight tensor is superior to others for FC2 layer (better viewed together with Table \ref{fc1} and Table \ref{fc2}). }
	\label{dimension_fc}    
\end{figure*}

We also conducted experiments about dimension adjustment of FC1 and FC2 layers in LeNet5. We conducted this experiments by changing the order of the weight tensor of FC1/FC2 layer while fixing the rest. The details of the FC1/FC2’s weight tensor with different orders are listed in the appendices part Table  \ref{fc1} and Table \ref{fc2}. The network performances under different settings are showing in Figure \ref{dimension_fc}. From Figure \ref{dimension_fc}, a fourth order weight tensor is optimal for FC1 layer while a fifth order weight tensor is superior to others for FC2 layer. The original orders for FC1 and FC2 are both two. Same as previous analysis,  the gray and yellow bars mean we did not use adaptive dimension adjustment on these two settings. Our ADA-tucker also got better results than Tucker without ADA on FC layers.

In summary, the optimal order of weight tensor varies for different layers. The results indicate that previous low-rank compression methods may impair network performance by constraining a fixed form of decomposition. The flexibility of ADA-Tucker enables networks to adaptively adjust the dimensions and orders of weight tensor when being compressed, thus achieving better performance.

The ADA-Tucker algorithm is summarized in Alg. \ref{algorithm}.

\begin{algorithm}[!ht]
	\caption{ADA-Tucker Algorithm}
	\begin{algorithmic}
		\label{algorithm}
		\STATE $\mathbf{Input}$: $\mathbf{X}$, $\mathbf{Y}$: training data and labels.\\
		\STATE $\mathbf{Input}$: $\{n_1^{(i)}, n_2^{(i)}, ..., n_{d_c}^{(i)}: 1 \leq i\leq l\}$: $n_{d_c}^{(i)}$ is the dimension of $d_c$ mode of the $i$-th layer's reshaped weight tensor, which is denoted by adaptive dimension adjustment mechanism.\\
		\STATE $\mathbf{Output}$: $\{\boldsymbol{\mathcal{C}}^{(i)},\mathbf{M}^{(i)}_1,\mathbf{M}^{(i)}_2,...,\mathbf{M}^{(i)}_{i_{d_c}}:1 \leq i\leq l\}$: the core tensors and transformation matrices for every layer. \\
		
		\STATE Adaptive dimension adjustment: based on the input $\{n_1^{(i)}, n_2^{(i)}, ..., n_{d_c}^{(i)}: 1 \leq i\leq l\}$, construct $\boldsymbol{\tilde{\mathcal{W}}}^{(i)}$ from $\boldsymbol{\mathcal{W}}^{(i)}$, define  $\boldsymbol{\mathcal{C}}^{(i)}$ and $\mathbf{M}^{(i)}_j$, $1\leq i\leq l, 1\leq j\leq i_{d_c}$.\\
		\STATE$\mathbf{for}$ number of  training iterations $\mathbf{do}$\\
		
		\STATE $\quad$Choose a minibatch of network input from $\mathbf{X}$.\\

		\STATE $\quad$$\mathbf{for}$ $i=1,2,3,...,l$ $\mathbf{do}$
		\STATE $\quad\quad$Use Eq.~\eqref{tucker_any} and reshape function to rebuild $\boldsymbol{\mathcal{W}}^{(i)}$, use $\boldsymbol{\mathcal{W}}^{(i)}$ to get the output of the $i$-th layer.\\
		\STATE $\quad\mathbf{end}\ \mathbf{for}$\\
		\STATE $\quad$Compute the loss function $L$.\\
		\STATE $\quad$$\mathbf{for}$ $i=l,l-1,l-2,...,1$ $\mathbf{do}$
		\STATE $\quad\quad$Follow traditional backward propagation to get $\frac{\partial L}{\partial \boldsymbol{\mathcal{W}}^{(i)}}$ and compute $\frac{\partial L}{\partial {\boldsymbol{\mathcal{\tilde{W}}}}^{(i)}}$ from $\frac{\partial L}{\partial \boldsymbol{\mathcal{W}}^{(i)}}$.\\
		\STATE $\quad\quad$$\mathbf{for}$ $j=1,2,3,...,i_{d_c}$ $\mathbf{do}$
		\STATE $\quad\quad\quad$ Use Eq.~\eqref{dm} to compute $\frac{\partial L}{\partial \boldsymbol{M}^{(i)}_j}$, then update $\boldsymbol{M}^{(i)}_j$.
		\STATE $\quad\quad$$\mathbf{end\ for}$
		\STATE $\quad\quad$ Use Eq.~\eqref{dc} to compute $\frac{\partial L}{\partial \boldsymbol{C}^{(i)}_{(1)}}$, use Eq.~\eqref{dc_reshape} to construct $\frac{\partial L}{\partial \boldsymbol{\mathcal{C}}^{(i)}}$, then update $\boldsymbol{\mathcal{C}}^{(i)}$.
		\STATE $\quad\mathbf{end}\ \mathbf{for}$
		\STATE$\mathbf{end}\ \mathbf{for}$
	\end{algorithmic}
\end{algorithm}

\subsubsection{Influence of Dimension Eveness of Core Tensor}\label{ada_shape} 

\begin{figure*}[t]
	\centering
	\includegraphics[width=0.8\linewidth]{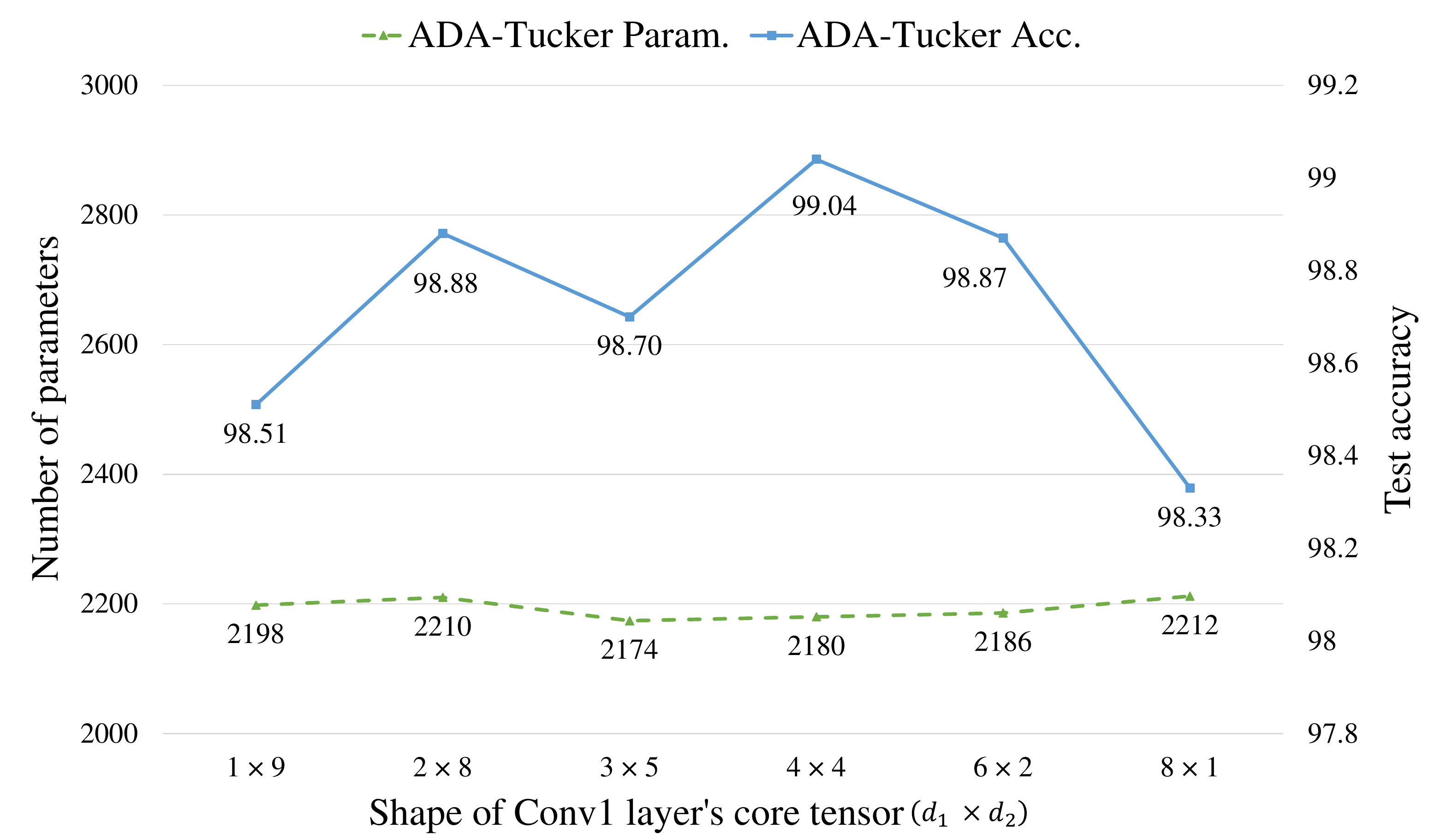}
	\caption{Influence of dimension eveness of core tensor. Experiments are conducted by gradually increasing the aspect ratio of the second order core tensor of Conv1 (first convolutional) layer ($25 \times 20$) in LeNet-5 while fixing the rest of the network architecture. We control the variation of each model's weight number within a negligible range and test their classification accuracy on MNIST with aspect ratio ranging from $1:9$ to $8:1$ (better viewed together with Table \ref{dim}).}
	\label{shape_selection} 
\end{figure*}

To explore the influence of dimension eveness of core tensor, we change the shape of core tensor and record accuracy of each network after training them from scratch. The details of this experiment settings please refer to Table \ref{dim} in appendices part. As is shown in Figure \ref{shape_selection}, the network with square core tensors performs better than the one with other core shapes. Specifically, as the difference between the two dimensions grows larger, i.e., when the core becomes less `square', the network's accuracy decreases accordingly. We speculate that the mechanism behind is to evenly distribute weights across different dimensions.

Here we give a more clear summary for the adaptive dimension adjustment mechanism based on the experiments on dimension adjustment of Conv layers, dimension adjustment of FC layers and influence of dimension eveness of core tensor: The mechanism is somewhat like factorization of the number of weights for a specific layer. The number of factors is equal to the order after the adaptive dimension adjustment mechanism. From the experiments of dimension adjustment of Conv layers (Figure \ref{dimension_conv} for performances of all settings. Table \ref{conv1} for the detail of Conv1’s weight tensor and Table \ref{conv2} for the detail of Conv2’s weight tensor) and FC layers (Figure \ref{dimension_fc} for performances of all settings. Table \ref{fc1} for the
detail of FC1’s weight tensor and Table \ref{fc2} for the detail of FC2’s weight tensor), we found that if we make factors’ values more similar with each other (balanced dimensions), the performance is better. The factors’ value should not be too big, otherwise it will cost vast storage for transformation matrices and lose much information (Performance degradation in FC1 layer when the order of core tenor is two). From the experiments of Influence of dimension eveness of core tensor, we learn that if the reshape tensor is balanced with proper order, the core tensor with hypercube shape will have the best performance.

\subsection{CP is a special case of Tucker and Tucker is a special case of ADA-Tucker}

Suppose now we have a $d$-dimensional tensor $\boldsymbol{\mathcal{W}}$ of size $n_1 \times n_2 \times... \times n_d$ and a core tensor $\boldsymbol{\mathcal{C}}$ of size $k_1 \times k_2 \times... \times k_d$, Tucker decomposition has the following form:
\begin{equation}
\begin{aligned}
\boldsymbol{\mathcal{W}} &\approx \boldsymbol{\mathcal{C}}\times_1 \boldsymbol{M}_1 \times_2 \boldsymbol{M}_2 \times_3 ... \times_d \boldsymbol{M}_d\\
&= \sum_{i_1=1}^{k_1}\sum_{i_2=1}^{k_2}...\sum_{i_d=1}^{k_d}\boldsymbol{\mathcal{C}}_{i_1i_2...i_d}\boldsymbol{m}_{i_1}^{1}\otimes\boldsymbol{m}_{i_2}^{2}\otimes...\otimes\boldsymbol{m}_{i_d}^{d},
\end{aligned}
\end{equation}
where $m_{i}^j$ means the $i$-th column of matrix $\boldsymbol{M}_j$. While CP-decomposition has the following form:
\vspace{-3pt}
\begin{equation}
\begin{aligned}
\boldsymbol{\mathcal{W}} &\approx  \sum_{i=1}^{r}\lambda_{i}\boldsymbol{m}_{i}^{1}\otimes\boldsymbol{m}_{i}^{2}\otimes...\otimes\boldsymbol{m}_{i}^{d}.
\end{aligned}
\end{equation}

In the case of the core tensor being a hypercube, if its elements are nonzero when $i_j=i,\forall j\in\{1,2,3...,d\}$ and are zero otherwise, then Tucker degenerates to CP. The fact that CP is a special case of Tucker indicates that Tucker is more powerful than CP. In fact, Tucker encodes much more compact representation as its core tensor is denser and smaller-sized, using the mechanism of Adaptive Dimension Adjustment. It is obvious to learn that ADA-Tucker degenerates to Tucker without using the mechanism of Adaptive Dimension Adjustment. Empirically, the following experimental evidence is provided for detailed comparisons for these three methods.

\subsection{Shared Core ADA-Tucker}\label{scada}

With input data passing serially through each layer in a network, we believe that there exist some correspondence and invariance in weights across different layers. Concretely, we think that a weight tensor preserve two kind of information, namely, the first kind of information tries to construct some transformations to extract global features (encode the same object at different layers and different scales) and the second kind of information tries to construct some transformations to extract local specific features for different layers and different scales. This conjecture indicates that there may exist shared information for transformation in functions expressed as a $d_c$-mode product between core tensors and transformation matrices across layers. We assume that the layer-invariant information lies in the core tensor, as it has the majority of weights, while the transformation matrices are responsible for the layer-specific mapping. Therefore, as illustrated in Figure \ref{tucker_illustration_scada}, we devise a so-called SCADA-Tucker, where all layers of a network share one common core tensor, thus achieving higher compression ratio.

Suppose that the network has $l$ layers. Based on the description above, we need one core tensor and $\sum_{i=1}^{l}d_i$ transformation matrices, where $d_i$ represents the order of core tensor for the $i$-th layer. With SCADA-Tucker, the model contains $l$ core tensors and $ld$ transformation matrices ($d=\max \{d_i\}, i=1,2,...,l$). We can set a specific transformation matrix $\mathbf{M}_{j}^{(i)} \in \mathbb{R}^{1\times k_j^{(i)}} (j=1,2,\dots,d)$ if the reshaped weight tensor of the $i$-th layer has a lower order than the shared core tensor. The forward propagation is: 

\begin{equation}\label{tucker_share}
\begin{aligned}
&{\boldsymbol{\mathcal{\tilde{W}}}}^{(i)}\approx\boldsymbol{\mathcal{C}}\times_1 \boldsymbol{M}_1^{(i)} \times_2 \boldsymbol{M}_2^{(i)}\times_3 ... \times_d \boldsymbol{M}_d^{(i)},\\ &  \boldsymbol{\mathcal{W}}^{(i)} = {\rm{reshape}}\left({\boldsymbol{\mathcal{\tilde{W}}}}^{(i)}\right).
\end{aligned}\end{equation}

\begin{figure*}[t]
	\centering
	\includegraphics[width=1\linewidth]{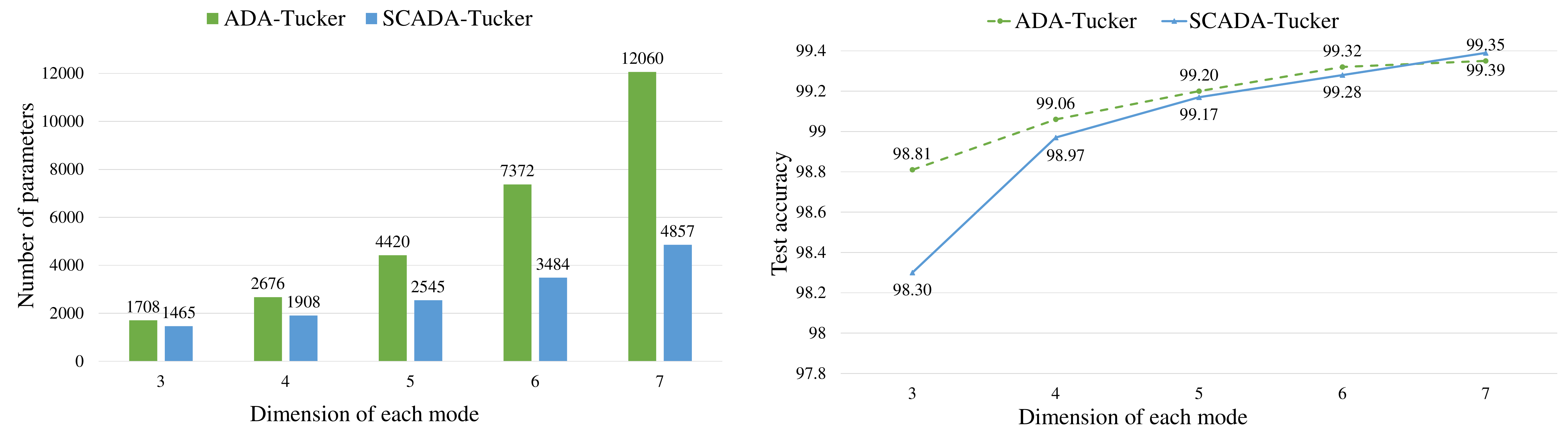}
	\caption{SCADA-Tucker vs. ADA-Tucker comparing the number of weights and network's performance (better viewed together with Table \ref{scada_ada}).}
	\label{ADA_SCADA}   
\end{figure*}
The backpropagation of $ld$ transformation matrices are the same as that in the ADA-Tucker model. The major difference lies in the gradient w.r.t. the core tensor. We compute it as:

\begin{equation}\label{share}
\frac{\partial L}{\partial \boldsymbol{\mathcal{C}}}= \sum_{i=1}^{l}\frac{\partial L}{\partial \boldsymbol{\mathcal{C}^{(i)}}}.
\end{equation}
We present here a detailed property comparison between SCADA-Tucker and ADA-Tucker. We use LeNet-5 and set the core tensor of each layer as a fourth order tensor of the same size. We examine the performance of LeNet-5 using two compression methods by changing the size of core tensor only while fixing the rest hyperparameters. The details of ADA-Tucker and SCADA-Tucker settings for this experiments are in Table \ref{scada_ada} of appendices part. From the results in Figure \ref{ADA_SCADA}, we can see that under the same parameter setting, SCADA-Tucker is able to significantly reduce the number of weight in the network  with only minor performance degradation compared to ADA-Tucker. It is because core tensors generally account for a major proportion of the total number of weights. When the dimension of each mode increases to seven, SCADA-Tucker even achieves an accuracy slightly higher than that of ADA-Tucker. Note that the number of weight in SCADA-Tucker is less than one half of that in ADA-Tucker under the same setting.

\begin{figure*}[t]
	\centering
	\includegraphics[width=0.95\linewidth]{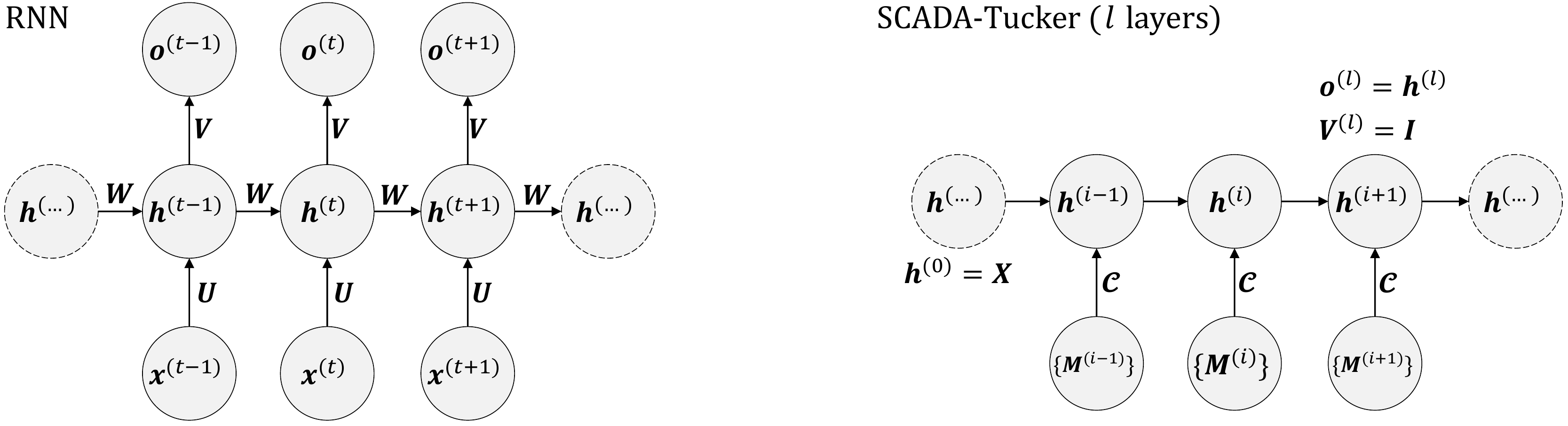}
	\caption{The comparison of forward propagations for RNN (left) and SCADA-Tucker (right).}
	\label{SCADA_rnn}   
\end{figure*}
An alternative perspective is to view SCADA-Tucker as a module with properties analogous to recurrent neural networks (RNN). The  comparison of forward propagations for these two models can be seen in Figure \ref{SCADA_rnn}. We all know that an RNN can be rolled out as a serial network with shared weights and it captures the temporal relations among sequential inputs. Thus with part of weights shared across layers, SCADA-Tucker can be regarded as an architecture in a recurrent style. Concretely, the forward propagation of RNN can be represented as $\boldsymbol{h}^{(t)}=\sigma(\boldsymbol{U}\boldsymbol{x}^{(t)}+\boldsymbol{W}\boldsymbol{h}^{(t-1)}+\boldsymbol{b})$, where $\boldsymbol{x}^{(t)}$ is the input at time step $t$, $\boldsymbol{h}^{(t)}$ is the hidden state at time step $t$,  $\boldsymbol{b}$ is the bias term, $\sigma(\cdot)$ is the activation function and $\boldsymbol{U}, \boldsymbol{W}$ are the transformation matrices shared by all time steps. In comparison, the forward propagation of SCADA-Tucker can be represented as

\begin{equation}
\boldsymbol{h}^{(i)}=\sigma(\boldsymbol{\mathcal{C}}\prod_{j}\times_j\boldsymbol{M}_j^{(i)}\boldsymbol{h}^{(i-1)}+\boldsymbol{b}), i = 1,2,3,...,l.
\end{equation}

The ``input'' of each layer is a series of transformation matrices $\{\boldsymbol{M}\}$ and the core tensor $\boldsymbol{\mathcal{C}}$ is the share parameter for all layers. 

Moreover, SCADA-Tucker creatively connects weights of different layers, enabling direct gradient flow from the loss function to earlier layers. Some modern architectures of neural networks such as ResNet \cite{he2016deep}, DenseNet \cite{huang2016densely} and CliqueNet \cite{yang2018convolutional} benefit from the direct gradient flow from the loss function to earlier layers and have achieved great success. \textit{Such a parameter reuse mechanism also addresses redundancy in common parameterized functions shared across layers. Note that none of other compression methods involve dimension adjustment and sharing parameters among different layers}, which are crucial to the compression performance.

Therefore,  SCADA-Tucker has the potential to pass and share some critical and common information across layers, which cannot be achieved by ADA-Tucker. Finally, we regard SCADA-Tucker as a promising compression method for high-ratio network compression with a negligible sacrifice in network performance.

\section{Compression Ratio Analysis}\label{compression_analysis}

\subsection{Raw Compression Ratio Analysis}

Suppose that the network has $l$ layers. Let $\boldsymbol{\mathcal{\tilde{W}}}^{(i)}\in \mathbb{R}^{n^{(i)}_1\times n^{(i)}_2 \cdot\cdot\cdot  \times n^{(i)}_{d_i}}$ and $\boldsymbol{\mathcal{C}}^{(i)}\in \mathbb{R}^{k^{(i)}_1\times k^{(i)}_2 \cdot\cdot\cdot  \times k^{(i)}_{d_i}}$ be the reshaped weight tensor and core tensor of the $i$-th layer, respectively. Obviously, $\boldsymbol{\mathcal{\tilde{W}}}^{(i)}$ has the same number of weights as $\boldsymbol{\mathcal{W}}^{(i)}$. Then the compression ratio of ADA-Tucker is:
\vspace{-3pt}
\begin{equation}\label{c_a}
\begin{aligned}
r_{A}&= \frac{\sum_{i=1}^{l}\prod_{j=1}^{d_i}n^{(i)}_j}{\sum_{i=1}^{l}\prod_{j=1}^{d_i}k^{(i)}_j+\sum_{i=1}^{l}\sum_{j=1}^{d_i}n^{(i)}_jk^{(i)}_j}\\
&\approx\frac{\sum_{i=1}^{l}\prod_{j=1}^{d_i}n^{(i)}_j}{\sum_{i=1}^{l}\prod_{j=1}^{d_i}k^{(i)}_j}.
\end{aligned}
\end{equation}

For SCADA-Tucker, all layers share the same core tensor with order $d$, i.e,\  $d=d_i, i=1,2,\cdots,l$. Then its compression ratio is:
\vspace{-3pt}
\begin{equation}\label{c_sc}
\begin{aligned}
r_{SC}&= \frac{\sum_{i=1}^{l}\prod_{j=1}^{d}n^{(i)}_j}{\prod_{j=1}^{d}k_j+\sum_{i=1}^{l}\sum_{j=1}^{d}n^{(i)}_jk_j}\\
&\approx\frac{\sum_{i=1}^{l}\prod_{j=1}^{d}n^{(i)}_j}{\prod_{j=1}^{d}k_j}\geqslant r_A.
\end{aligned}
\end{equation}

\subsection{Further Compression by Quantization}
\begin{figure*}[t]
	\centering
	\includegraphics[width=1\linewidth]{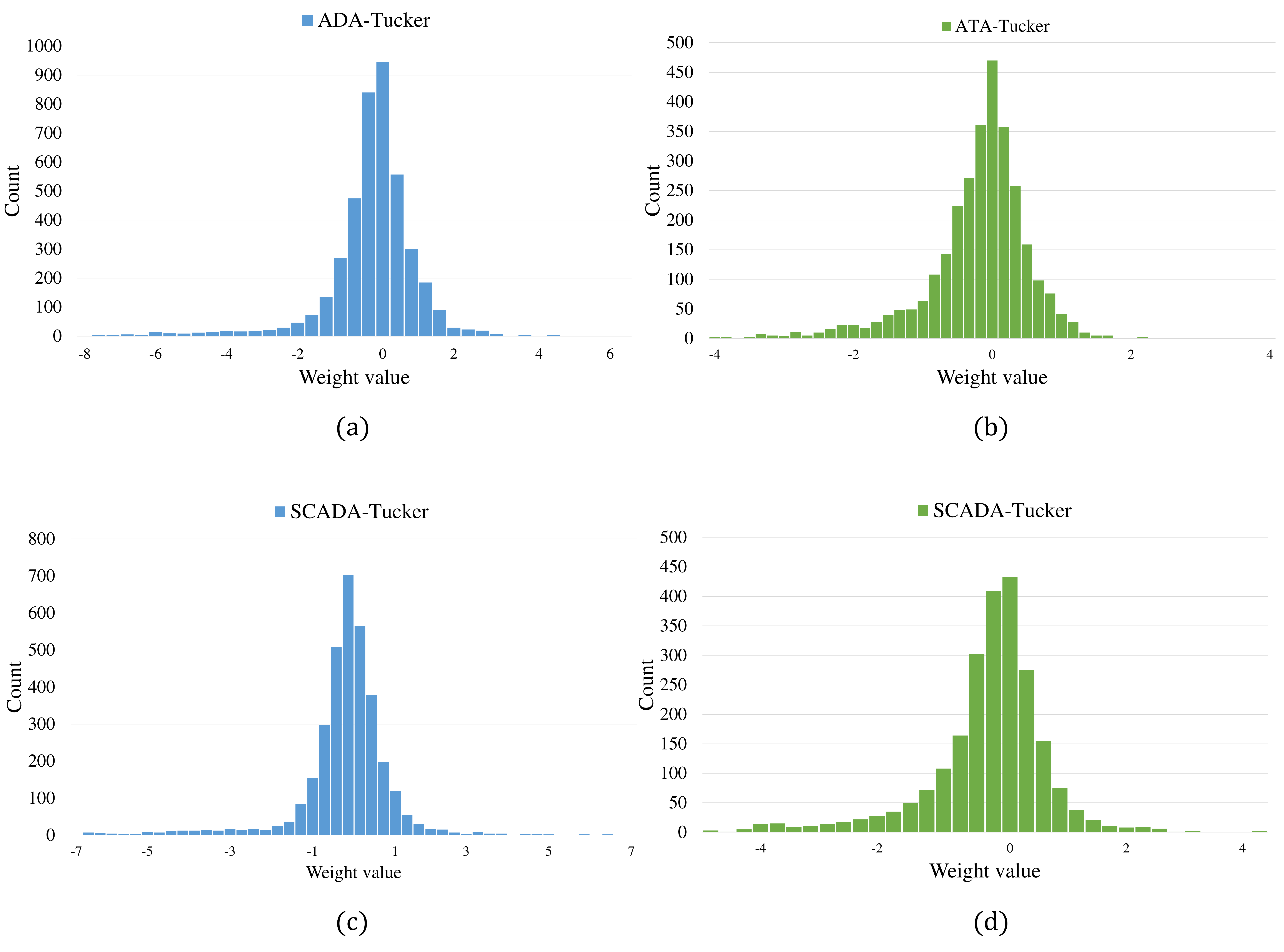}
	\caption{Weight distribution after compressing by our methods:  (a) ADA-Tucker on LeNet-300. (b) ADA-Tucker on LeNet-5. (c) SCADA-Tucker on LeNet-300. (d) SCADA-Tucker on LeNet-5. }
	\label{distribution}
\end{figure*}

After being compressed by ADA-Tucker and SCADA-Tucker, the weight distribution is close to Laplacian distribution. Since almost all weights are in the range of $[-3,3]$ (Figure \ref{distribution}), we can use pruning and quantization to compress these weights further following Han et al. \cite{han2015learning}. In our experiments, we integrate the following quantization into our model:
\begin{equation}\label{quantization}
w_q = -b + \left\lfloor\frac{(\max(-b,\min(w,b))+b))Q}{2b}\right\rfloor\cdot\frac{2b}{Q},
\end{equation}
where $Q$ represents the number of clusters, $b$ represents the max-min bound of quantization and $\lfloor x \rfloor$ is the floor function. Since weights are originally stored in the float32 format ($32$ bits), our compression ratio can be further increased by this quantization trick. After quantization, we utilize Huffman coding to compress it further. Suppose that the average length of Huffman coding is $\bar{a}$, we compute the final compression ratio by:
\vspace{-0.1cm}
\begin{equation}\label{c_qh}
r_{A+QH}=\frac{32r_{A}}{\bar{a}},\ \  r_{SC+QH}=\frac{32r_{SC}}{\bar{a}}.
\end{equation}

\section{Experimental Results}\label{experiment}

In this section, we experimentally analyze the proposed methods. We use Adam \cite{kingma2014adam} as our optimizer. The starting learning rate is set as $0.01$, $0.003$, $0.003$ and $0.01$ for MNIST, CIFAR-10, SVHN and ImageNet, respectively. After every $10\sim20$ epochs, the learning rate is divided by $3$. We choose $256$ as the batch size for most experiments. For initializers of core tensors and transformation matrices, we have experimented with the glorot initializer \cite{glorot2010understanding}, the Kaiming initializer \cite{he2015delving} and HOOI \cite{de2000best} to solve the decomposition from the original weight tensors. These three methods have similar performances.  Note that for the following experiments, little time is spent on fine-tuning our model.

\vspace{-3pt}
\subsection{MNIST}\label{mnist_experiment}

MNIST is a database of handwritten digits with $60,000$ training images and $10,000$ testing images. It is widely used to evaluate machine learning algorithms. Same as DC \cite{han2015deep}, DNS \cite{guo2016dynamic} and SWS \cite{ullrich2017soft}, we test our methods on two classical networks: LeNet-5 and LeNet-300-100. 
\begin{table*}[t]
	\centering
	\begin{tabular}{ccc|c|c|c|c|c}
		\hline
		\multirow{2}{*}{Network}   & \multirow{2}{*}{Methods}  & \multirow{2}{*}{\#(Param.)} &\multicolumn{3}{c}{Test Error Rate[\%]}  & \multicolumn{2}{|c}{CR} \\
		\cline{4-8}
		& & & Org. & Raw & +QH & Raw& +QH \\ 
		\hline \hline
		\multirow{5}{*}{\begin{tabular}[c]{@{}c@{}}LeNet-\\ 300-100\end{tabular}} & DC \cite{han2015deep} & $21.4$K  & $ 1.64$&$1.57$   &$1.58$ &$<12$ &$40$ \\
		& DNS \cite{guo2016dynamic}  & $4.8$K &$ 2.28$&$1.99$& -    &$< 56$ &-  \\
		& SWS \cite{ullrich2017soft} & $4.3$K & $ 1.89$&-&$1.94$  &$<62$ &$64$  \\
		& ADA-Tucker & $4.1$K   & $ 1.89$&$1.88$ &$1.91$ &$=65$ &$\mathbf{233}$  \\
		& SCADA-Tucker   & $3.4$K &$ 1.89$ &$ 2.11$&$ 2.26$ &$=78$&$\mathbf{321}$ \\
		\hline \hline
		\multirow{5}{*}{LeNet-5} & DC \cite{han2015deep} & $34.5$K  & $ 0.80$&$0.77$&$0.74$   &$<13$ &$39$  \\
		
		&  DNS \cite{guo2016dynamic} & $4.0$K  & $ 0.91$&$0.91$ &- &$<108$ &-  \\
		& SWS \cite{ullrich2017soft} & $2.2$K & $ 0.88$&-&$0.97$  &$<196$ &$162$  \\
		& ADA-Tucker & $2.6$K   & $ 0.88$&$0.84$ &$0.94$ &$=166$ &$\mathbf{691}$  \\
		& SCADA-Tucker   & $2.3$K  & $ 0.88$&$0.94$&$1.18$ &$=185$ &$\mathbf{757}$  \\
		\cline{1-8}
	\end{tabular}
	%       }
	\captionof{table}{Compression results on LeNet-5 and LeNet-300-100. +QH: adding quantization and Huffman coding after utilizing these methods. \#(Param.) means the total number of parameters of these methods. CR represents compression ratio. }
	\label{mnist}
\end{table*}

The raw compression ratios of ADA-Tucker and SCADA-Tucker in Table \ref{mnist} are computed by Eq. \eqref{c_a} and Eq. \eqref{c_sc}, respectively. The compression ratio of +QH is computed by Eq. \eqref{c_qh}. Because all these methods \cite{guo2016dynamic,han2015learning,ullrich2017soft} in Table \ref{mnist} \footnote{ We compare our models with state of the art in 2015, 2016 and 2017 for compressing LeNet-5 and LeNet-300. HashedNet \cite{chen2015compressing} does not appear in Table \ref{mnist} because it used a network different from LeNet-5 or LeNet-300 and thus cannot be compared with other methods. Since methods in Table \ref{mnist} conducted experiments on MNIST but not on CIFAR-10 or SVHN, these methods are not shown in Table \ref{cifar_svhn}.} need to record the indices of nonzero elements, their actual compression ratios are smaller than the calculated results. \textit{Our methods do not need to record the indices}, so our actual compression ratios are equal to the calculated results, suggesting that our model has the highest compression ratio even if it has the same number of weight with the methods mentioned above. We set $Q=512$ and $b=3$ during the quantization process of LeNet-5 and get the final compression ratio of $\mathbf{691\times}$ with $0.94\%$ error rate. For LeNet-300-100, we set $Q=1,500$ and $b=5$ to achieve a $\mathbf{233\times}$ compression ratio and the final error rate is $1.91\%$. The value of $b$ can be adjusted according to the distribution of weights after ADA-Tucker/SCADA-Tucker compression.

Tensor Train decomposition (TT) \cite{novikov2015tensorizing} is similar to Tucker decomposition in that they both involve a product of matrices. With Tucker, the number of parameters can be further reduced by sharing core tensor, which cannot be achieved by TT. Moreover, we chose Tucker because TT has to use enormously-sized matrices to exactly represent a tensor when its order is greater than three. Thus compressing with TT significantly may cause huge approximation error. Using Tucker helps strike a better balance between compression ratio and recognition accuracy. More importantly, TT can only be applied to FC layers despite the fact that Conv layers are more crucial than FC layers to achieve top performance for most of the current deep learning tasks. In contrast, our model with Tucker decomposition is able to adjust the order and dimension of tensors in \textit{both FC and Conv layers}. Still, for a closer examination, here we provide results of two models in all-FC-layer network for better reference. For MNIST we use the same network architecture as \cite{novikov2015tensorizing} and get $98.13\%$ test accuracy with $6,824$ parameters, while TT gets $98.1\%$ test accuracy with $7,698$ parameters. This again proves the strength of our methods in compressing network and preserving information.

\begin{table*}[!ht]
	\small
	\centering
	\setlength{\tabcolsep}{0.7mm}{
		\begin{tabular}{cc|c|c|c|c|c|c|c|c}
			\hline
			\multirow{1}{*}{Dataset}      & \multirow{1}{*}{CNN-ref}     &\multicolumn{2}{|c}{LRD} &\multicolumn{2}{|c}{HashedNet}  &\multicolumn{2}{|c|}{FreshNet} & ADA-Tucker & SCADA-Tucker\\
			\cline{1-10}
			\multicolumn{2}{c|}{CR} &  $16 \times$ & $64 \times$ &  $16 \times$ & $64 \times$ &  $16 \times$ & $64 \times$ & $64\times$& $73\times$\\
			\hline \hline
			CIFAR-10 & $14.37$  & $23.23$&$34.35$&$24.70$&$43.08$  & $21.42$&$30.79$ & $\mathbf{17.97}$ & $\mathbf{20.27}$  \\
			\cline{1-10}
			SVHN & $3.69$  & $10.67$&$22.32$ & $9.00$&$23.31$  & $8.01$&$18.37$ & $\mathbf{4.41}$ & $\mathbf{3.92}$  \\
			\hline
			
	\end{tabular}}
	\caption{ Test error rates (in \%) with compression ratio at $16\times$ and $64\times$ for LRD \cite{denil2013predicting}, HashedNet \cite{chen2015compressing}, FreshNet \cite{chen2016compressing} and ours. CR represents compression ratio.}
	\label{cifar_svhn}
\end{table*}

\subsection{SVHN and CIFAR-10}\label{svhn_cifar}

To prove the generalization ability of our methods, we also conduct experiments on SVHN and CIFAR-10 datasets. The SVHN dataset is a large collection of digits cropped from real-world scenes, consisting of $604,388$ training images and $26,032$ testing images. The CIFAR-10 dataset contains $60,000$ images of $32 \times 32$ pixels with three color channels. With the same network architectures, our compressed models significantly outperform \cite{denil2013predicting,chen2015compressing,chen2016compressing} in terms of both compression ratio and classification accuracy. The details of network architecture are listed in Table \ref{cifar_net} and the setting for ADA-Tucker are showing in Table \ref{cifar_ada} of appendices part. On the CIFAR-10 dataset, ADA-Tucker has a higher accuracy with a compression ratio lower than SCADA-Tucker as expected. However, on the SVHN dataset, SCADA-Tucker surprisingly preforms much better than ADA-Tucker. Specifically, SCADA-Tucker compresses the original network by $73 \times$ with $0.23\%$ accuracy drop, while ADA-Tucker compresses it by $64 \times$ with $0.72\%$ accuracy drop.

\vspace{-3pt}
\subsection{ILSVRC12}

In this subsection, we empirically compare the performances of CP, Tucker and ADA-Tucker on ILSVRC12 dataset.

\begin{figure*}
	\centering
	\includegraphics[width=0.85\textwidth]{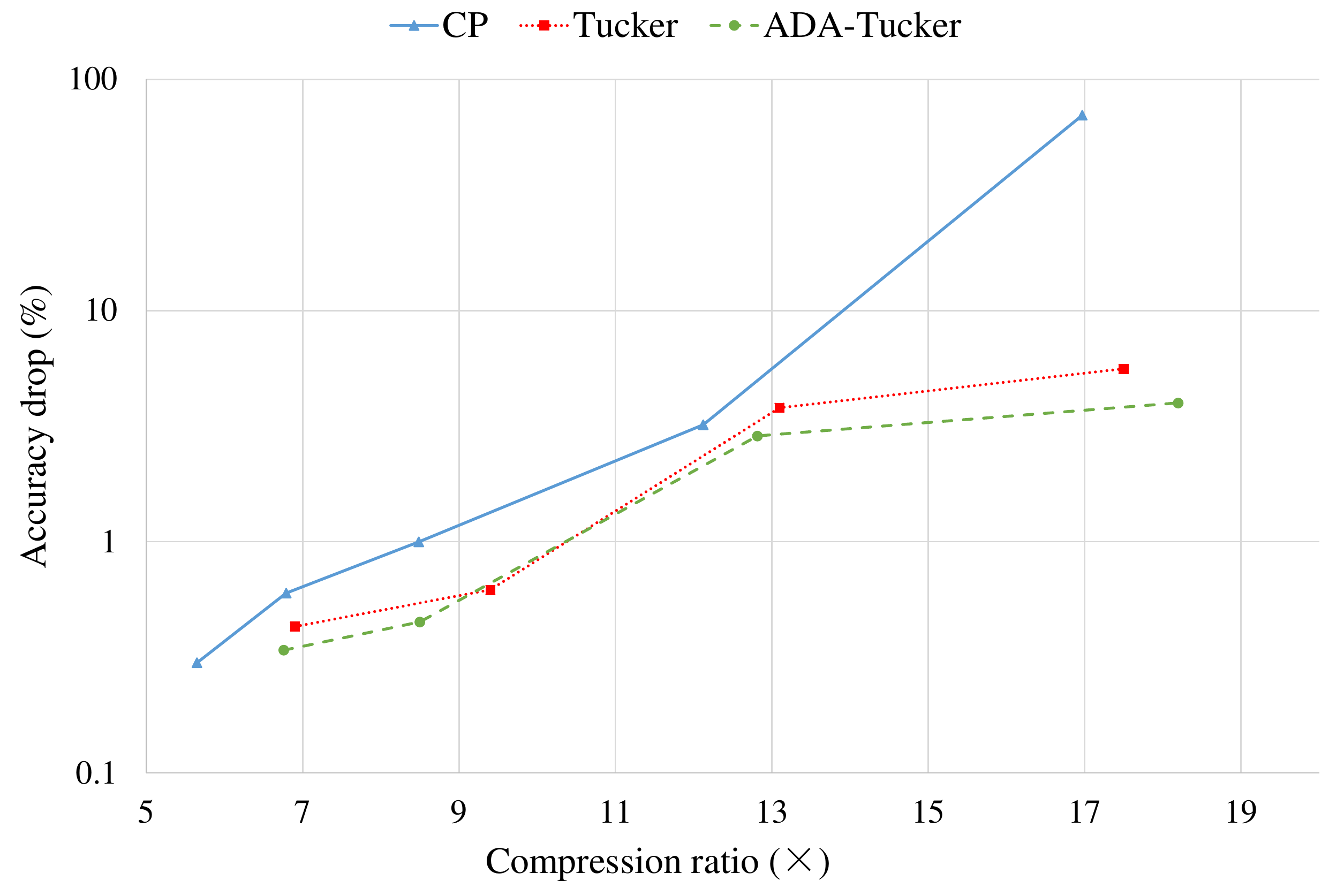}
	\vspace{-0.1cm}
	\label{cp_vs_tucker}
	\caption{Comparison of ADA-Tucker, Tucker-decomposition and CP-decomposition on ImageNet. (logarithmic coordinates)}
	\label{ada_vs_cp}
\end{figure*}
To prove this work preserves more information and easier compress networks compared with CP-decomposition and Tucker-decomposition, we follow the ILSVRC12 experiment in \cite{lebedev2014speeding}. We also compress the second convolutional layer of AlexNet \cite{krizhevsky2012imagenet}. As a baseline, we use a pre-trained AlexNet model shipped with Pytorch, which achieves a top-5 accuracy of $79.59\%$. Following \cite{lebedev2014speeding}, models are evaluated by test accuracy drop when increasing compression ratio. Experimental results in Figure \ref{ada_vs_cp} shows that our methods have less accuracy drop at the same compression ratio. The gap between our method and CP-decomposition becomes larger as the compression ratio goes higher. The same experimental phenomenon also appeared when we compared our method with Tucker-decomposition. Concretely, at the same compression ratio equal to 18, the accuracy drop of our method is less than $4\%$, while the CP-decomposition method drops about $70\%$ and Tucker-decomposition method drops about $6\%$. This result suggests that our method has a better capacity to preserve more information than CP and easier compress networks than Tucker.

\subsection{Modern networks}\label{modern_net}

\begin{table}[t]
	\centering
	\setlength{\tabcolsep}{1.8mm}{
		\begin{tabular}{cccccc}
			\hline
			Network            & \#(Param.) & Orig. Acc.(\%) & ADA-Tucker Acc.(\%) &    $\Delta$(Acc.)     & CR \\ \hline \hline
			ResNet-20          & 0.27M      & 91.25\%        & 90.97\%             & -0.28\% & \textbf{12} \\\hline
			WRN-28-10 & 36.5M      & 95.83\%        & 95.06\%             & -0.77\% & \textbf{58} \\\hline
	\end{tabular}}
	\caption{Compression results on ResNet-20 and WRN-28-10 on CIFAR-10 dataset. \#(Param.) means the total number of parameters of these methods. CR represents compression ratio.}
	\label{modern_network}
\end{table}

\begin{figure*}[t]
	\centering
	\includegraphics[width=1\textwidth]{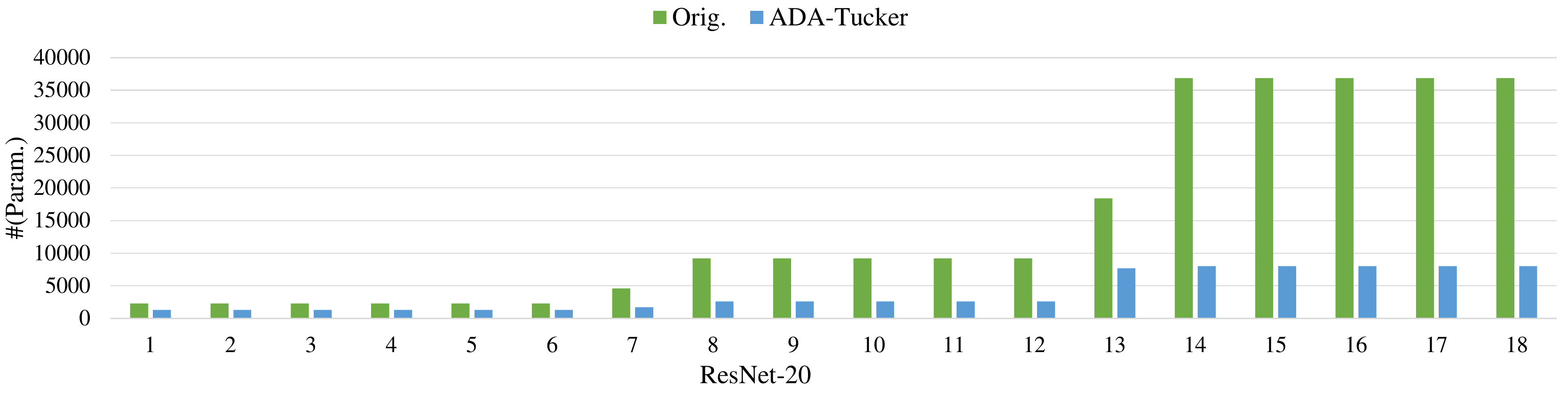}
	\vspace{-5mm}
	\caption{Parameters comparison of all convolutional layers in ResNet-20 (better viewed together with Table \ref{rnrn20}).}
	\label{resnet20}
\end{figure*}

\begin{figure*}[t]
	\centering
	\includegraphics[width=1\textwidth]{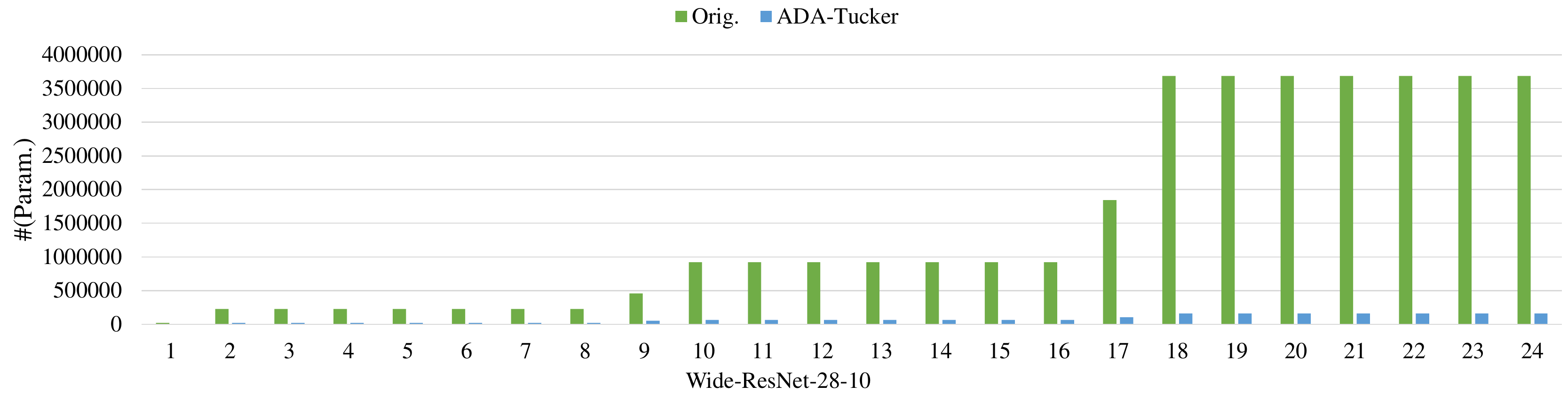}
	\caption{Parameters comparison of all convolutional layers in Wide-ResNet-28-10 (better viewed together with Table \ref{wrnwrn}).}
	\label{wrn2810}
\end{figure*}

Here we discuss a more recent work, ResNet \cite{he2016deep} and its variations Wide-ResNet \cite{zagoruyko2016wide, xie2017aggregated}. ResNet and its variations also have achieved promising preformances in numerous computer vision applications such as image classification, human face verification, object recognition, and object detection. It is very meaningful to be able to effectively compress these networks. 

We applied our ADA-Tucker on two representative networks ResNet-20 and Wide-ResNet-28-10 (WRN-28-10). This experiment was done on CIFAR-10 dataset. The details of ADA-Tucker for ResNet-20 and WRN-28-10 can be found in Table \ref{rnrn20} and Table \ref{wrnwrn} of appendices part, respectively. The compression results are listed in Table \ref{modern_network}. From Table \ref{modern_network}, ADA-tucker compressed ResNet-20 by $12\times$. Since the number of parameters of ResNet-20 is only about 0.27M, it is difficult to compress it further on CIFAR-10 dataset with negligible loss. The number of parameters of Wide ResNet-28-10 is about 36.5M, which is much bigger than ResNet-20's. Showing in Table \ref{modern_network}, our ADA-Tucker compressed WRN-28-10 by amazing 58 times without deteriorating its performances. We also plotted visualizations for parameters comparisons of all layers in terms of ResNet-20 (Figure \ref{resnet20}) and WRN-28-10 (Figure \ref{wrn2810}). The convincing results on these newly large networks suggest that the proposed method works well for modern CNN architectures.

\section{Conclusion}\label{conclusion}

In this paper, we demonstrate that deep neural networks can be better compressed using weight tensors with proper orders and balanced dimensions of modes without performance degradation. We also present two methods based on our demonstration, ADA-Tucker and SCADA-Tucker, for deep neural network compression. Unlike previous decomposition methods, our methods adaptively adjust the order of original weight tensors and the dimension of each mode before Tucker decomposition. We do not need to add new layers for implementing the Tucker decomposition as other methods do. The advantage of our methods over those involving the frequency domain and pruning is that we do not require recording the indices of nonzero elements. We demonstrate the superior compressing capacity of the proposed model: after applying quantization and Huffman coding, ADA-Tucker compresses LeNet-5 and LeNet-300-100 by $\mathbf{691\times}$ and $\mathbf{233\times}$, respectively, outperforming state-of-the-art methods. The experiments on CIFAR-10 and SVHN also show our models' overwhelming strength. The experiments on ImageNet indicate that Tucker decomposition combined with adaptive dimension adjustment has a great advantage over other decomposition-based methods especially at a large compression ratio.The convincing results on these newly large networks also suggest that the proposed method works well for modern CNN architectures. 

In the future, we will further investigate the mechanism behind our findings and summarize a detailed rule of thumb for determining the order of weight tensor  as well as dimensions of modes. Other research directions include combining this work with pruning techniques and exploiting its potential in accelerating computation and inference.

\section*{Acknowledgments}

This research is partially supported by National Basic Research Program of China (973 Program) (grant nos. 2015CB352303 and 2015CB352502), National Natural Science Foundation (NSF) of China (grant nos. 61231002, 61625301, 61671027 and 61731018), Qualcomm, and Microsoft Research Asia.

\bibliographystyle{apalike2}
\bibliography{ref}

\newpage
\begin{appendices}
	\section{Experiments settings}

	\begin{table}[h]
		\centering
		\renewcommand\arraystretch{1.1}
		\setlength{\tabcolsep}{0.5mm}{
			\begin{tabular}{c|c|c|c|c|c|c}
				\hline
				Conv1      & Orig.       & 2d       & 3d        & 4d         & 5d          & 6d           \\ \hline \hline
				Shape      & 20$\times$1$\times$5$\times$5    & 25$\times$20    & 20$\times$5$\times$5    & 20$\times$1$\times$5$\times$5   & 5$\times$5$\times$5$\times$2$\times$2   & 5$\times$5$\times$5$\times$2$\times$2$\times$1  \\ \hline
				Core       & -           & 5$\times$5      & 4$\times$4$\times$4     & 3$\times$3$\times$3$\times$3    & 2$\times$2$\times$2$\times$2$\times$2   & 2$\times$2$\times$2$\times$2$\times$2$\times$2  \\ \hline \hline
				Conv2      & \multicolumn{6}{c}{Orig.: 50$\times$20$\times$5$\times$5, reshape: 50$\times$25$\times$20, core size: 5$\times$5$\times$5}    \\
				FC1        & \multicolumn{6}{c}{Orig.: 800$\times$500, reshape: 40$\times$25$\times$20$\times$20, core size: 5$\times$5$\times$5$\times$5} \\
				FC2        & \multicolumn{6}{c}{Orig.: 500$\times$10, reshape: 25$\times$20$\times$10, core size: 5$\times$5$\times$5}       \\ \hline \hline
				\#(Param.) & 3230        & 2980     & 2914      & 2904       & 2810        & 2834        \\ \hline
		\end{tabular}}
		
		\caption{The details of Conv1 layer's dimension adjustment experiment setting (In Sec. \ref{ada_conv}). In this experiment, we fixed the rest layers of LeNet-5.}
		\label{conv1}
	\end{table}

\vspace{15pt}

	\begin{table}[h]
		\centering
		\small
		\renewcommand\arraystretch{1.1}
			\setlength{\tabcolsep}{0.3mm}{
				\begin{tabular}{c|c|c|c|c|c|c}
					\hline
					Conv1      & \multicolumn{6}{c}{Orig.: 20$\times$1$\times$5$\times$5, reshape: 25$\times$20, core size: 5$\times$5}          \\ \hline\hline
					Conv2      & Orig.      & 2d       & 3d        & 4d         & 5d           & 6d            \\ \hline
					Shape      & 50$\times$20$\times$5$\times$5  & 250$\times$100  & 50$\times$25$\times$20  & 50$\times$20$\times$5$\times$5  & 10$\times$10$\times$10$\times$5$\times$5 & 10$\times$10$\times$5$\times$5$\times$5$\times$2 \\ \hline
					Core       & -          & 20$\times$10    & 5$\times$5$\times$5     & 4$\times$4$\times$4$\times$4    & 3$\times$3$\times$3$\times$3$\times$3    & 2$\times$2$\times$2$\times$2$\times$2$\times$2   \\ \hline \hline
					FC1        & \multicolumn{6}{c}{Orig.: 800$\times$500, reshape: 40$\times$25$\times$20$\times$20, core size: 5$\times$5$\times$5$\times$5} \\
					FC2        & \multicolumn{6}{c}{Orig.: 500$\times$10, reshape: 25$\times$20$\times$10, core size: 5$\times$5$\times$5}       \\ \hline \hline
					\#(Param.) & 27380      & 8580     & 2980      & 2956       & 2743         & 2518          \\ \hline
		\end{tabular}}
		\caption{The details of Conv2 layer's dimension adjustment experiment setting (In Sec. \ref{ada_conv}). In this experiment, we fixed the rest layers of LeNet-5.}
		\label{conv2}
	\end{table}
	
	\vspace{15pt}

	\begin{table}[h]
		\centering
		\small
		\renewcommand\arraystretch{1.1}
			\setlength{\tabcolsep}{0.2mm}{
				\begin{tabular}{ c|c|c|c|c|c|c}
					\hline
					Conv1      & \multicolumn{6}{c}{Orig.: 20$\times$1$\times$5$\times$5, reshape: 25$\times$20, core size: 5$\times$5}         \\
					Conv2      & \multicolumn{6}{c}{Orig.: 50$\times$20$\times$5$\times$5, reshape: 50$\times$25$\times$20, core size: 5$\times$5$\times$5}   \\ \hline\hline
					FC1        & Orig.   & 2d      & 3d        & 4d          & 5d             & 6d            \\ \hline
					Shape      & 800$\times$500 & 800$\times$500 & 500$\times$25$\times$20 & 40$\times$25$\times$20$\times$20 & 25$\times$16$\times$10$\times$10$\times$10 & 25$\times$10$\times$8$\times$8$\times$5$\times$5 \\ \hline
					Core       & -       & 10$\times$10   & 5$\times$5$\times$5     & 5$\times$5$\times$5$\times$5     & 4$\times$4$\times$4$\times$4$\times$4      & 3$\times$3$\times$3$\times$3$\times$3     \\ \hline\hline
					FC2        & \multicolumn{6}{c}{Orig.: 500$\times$10, reshape: 25$\times$20$\times$10, core size: 5$\times$5$\times$5}      \\ \hline \hline
					\#(Param.) & 402K    & 14930   & 4680      & 2980        & 3138           & 2742          \\ \hline
		\end{tabular}}
		\caption{The details of FC1 layer's dimension adjustment experiment setting (In Sec. \ref{ada_fc}). In this experiment, we fixed the rest layers of LeNet-5.}
		\label{fc1}
	\end{table}

	\begin{table}[bp]
		\centering
		\small
		\renewcommand\arraystretch{1.1}
		\setlength{\tabcolsep}{1mm}{
			\begin{tabular}{c|c|c|c|c|c|c}
				\hline
				Conv1      & \multicolumn{6}{c}{Orig.: 20$\times$1$\times$5$\times$5, reshape: 25$\times$20, core size: 5$\times$5}       \\
				Conv2      & \multicolumn{6}{c}{Orig.: 50$\times$20$\times$5$\times$5, reshape: 50$\times$25$\times$20, core size: 5$\times$5$\times$5} \\
				FC1        & \multicolumn{6}{c}{Orig.: 500$\times$10, reshape: 25$\times$20$\times$10, core size: 5$\times$5$\times$5}    \\ \hline\hline
				FC2        & Orig.    & 2d       & 3d         & 4d          & 5d         & 6d           \\ \hline
				Shape      & 500$\times$10   & 500$\times$10   & 25$\times$20$\times$10   & 20$\times$10$\times$10$\times$5  & 8$\times$5$\times$5$\times$5$\times$5  & 5$\times$5$\times$5$\times$5$\times$4$\times$2  \\ \hline
				Core       & -        & 6$\times$6      & 5$\times$5$\times$5      & 4$\times$4$\times$4$\times$4     & 3$\times$3$\times$3$\times$3$\times$3  & 2$\times$2$\times$2$\times$2$\times$2$\times$2  \\ \hline\hline
				\#(Param.) & 7580     & 5676     & 2980       & 3016        & 2907       & 2700         \\ \hline
		\end{tabular}}
		\caption{The details of FC2 layer's dimension adjustment experiment setting (In Sec. \ref{ada_fc}). In this experiment, we fixed the rest layers of LeNet-5.}
		\label{fc2}
	\end{table}

	\begin{table}[h]
		\centering
		\renewcommand\arraystretch{1.1}
			\setlength{\tabcolsep}{5mm}{
				\begin{tabular}{c|c|c|c|c|c|c}
					\hline
					Conv1    & \multicolumn{6}{c}{20$\times$1$\times$5$\times$5}                                                 \\\hline
					Shape    & \multicolumn{6}{c}{25$\times$20}                                                    \\\hline
					Core     & 1$\times$9         & 2$\times$8         & 3$\times$5         & 4$\times$4        & 6$\times$2       & 8$\times$1       \\\hline\hline
					Conv2    & \multicolumn{6}{c}{Orig.: 50$\times$20$\times$5$\times$5, reshape: 50$\times$25$\times$20, core size: 4$\times$4$\times$4}    \\
					FC1      & \multicolumn{6}{c}{Orig.: 800$\times$500, reshape: 40$\times$25$\times$20$\times$20, core size: 4$\times$4$\times$4$\times$4} \\
					FC2      & \multicolumn{6}{c}{Orig.: 500$\times$10, reshape: 25$\times$20$\times$10, core size: 4$\times$4$\times$4}       \\\hline\hline
					\#(Param.) & 2198        & 2210        & 2175        & 2180       & 2186      & 2212\\\hline  
		\end{tabular}}
		\caption{The details of Conv1 layer's influence of dimension experiment setting (In Sec. \ref{ada_shape}). In this experiment, we fixed the rest layers of LeNet-5.}
		\label{dim}
	\end{table}
	
	\begin{table}[h]
		\centering
		\small
		\renewcommand\arraystretch{1.1}
		\setlength{\tabcolsep}{0.2mm}{
			\begin{tabular}{c|c|c|c|c|c}
				\hline
				& Conv1                    & Conv2                    & FC1                      & FC2                      & \#(Param.)     \\ \hline\hline
				Original                                                                            & 20$\times$1$\times$5$\times$5                 & 50$\times$20$\times$5$\times$5                & 800$\times$500                  & 500$\times$10                   & 431K           \\\hline
				Reshape                                                                          & 20$\times$1$\times$5$\times$5                 & 50$\times$20$\times$5$\times$5                & 40$\times$25$\times$20$\times$20              & 25$\times$20$\times$5$\times$2                & 431K           \\ \hline\hline
				\multirow{4}{*}{\begin{tabular}[c]{@{}c@{}}Transformation\\ Matrices\end{tabular}} & 20$\times$c                     & 50$\times$c                     & 40$\times$c                     & 25$\times$c                     & 135c           \\
				& 1$\times$c                      & 20$\times$c                     & 25$\times$c                     & 20$\times$c                     & 66c            \\
				& 5$\times$c                      & 5$\times$c                      & 20$\times$c                     & 5$\times$c                      & 35c            \\
				& 5$\times$c                      & 5$\times$c                      & 20$\times$c                     & 2$\times$c                      & 32c            \\ \hline
				Bias                                                                             & 20                       & 50                       & 500                      & 10                       & 0.58K          \\ \hline\hline
				ADA-Tucker core                                                                  & \multirow{1}{*}{c$\times$c$\times$c$\times$c} & \multirow{1}{*}{c$\times$c$\times$c$\times$c} & \multirow{1}{*}{c$\times$c$\times$c$\times$c} & \multirow{1}{*}{c$\times$c$\times$c$\times$c} & 4${\rm{c}}^4$            \\ \hline
				ADA-Tucker total                                                                 &    ${\rm{c}}^4$+31c+20                      & ${\rm{c}}^4$+80c+50                           &   ${\rm{c}}^4$+105c+500                         &               ${\rm{c}}^4$+52c+10            & $4{\rm{c}}^4$+268c+0.58K \\ \hline\hline
				SCADA-Tucker core                                                                & \multicolumn{4}{c|}{\multirow{1}{*}{c$\times$c$\times$c$\times$c}}                                                              & ${\rm{c}}^4$             \\ \hline
				SCADA-Tucker total                                                               & \multicolumn{4}{c|}{${\rm{c}}^4$+(31c+20)+(80c+50)+(105c+500)+(52c+10)}                                                                                      & ${\rm{c}}^4$+268c+0.58K  \\ \hline
		\end{tabular}}
		\caption{The details of SCADA-Tucker vs. ADA-Tucker experiment setting (In Sec. \ref{scada}). In this experiment, we fixed the size of transformation matrices for all layers. Since there are four layers, ATA-tucker has four core tensors. $c$ can be equal to $3,4,5,6,7$.}
		\label{scada_ada}
	\end{table}
	
	\begin{table}[h]
		\centering
		\renewcommand\arraystretch{1.1}
		\setlength{\tabcolsep}{1.2mm}{
			\caption{The detail of network architecture used in Sec. \ref{svhn_cifar}. The network architecture was referred to \cite{denil2013predicting,chen2015compressing,chen2016compressing}. C: Convolution. RL: ReLU. MP: Max-pooling. DO: Dropout. FC: Fully-connected.}
			\label{cifar_net}
			\centering
			\begin{tabular}{cc|ccccc|c}
				\hline
				Layer     &Operation &Input dim. &Inputs &Outputs &C size &MP size& \#(Param.)\\
				\hline\hline
				
				1&C,RL&$32\times32$&3&32&$5\times5$&-&2K\\
				2&C,MP,DO,RL&$32\times32$&32&64&$5\times5$&$2\times2$&51K\\
				3&C,RL&$16\times16$&64&64&$5\times5$&-&102K\\
				4&C,MP,DO,RL&$16\times16$&64&128&$5\times5$&$2\times2$&205K\\
				5&C,MP,DO,RL&$8\times8$&128&256&$5\times5$&$2\times2$&819K\\
				6&FC,Softmax&-&4096&10&-&-&40K\\
				\hline
		\end{tabular}}
		
	\end{table}

	\begin{table}[h]
		\centering
		\renewcommand\arraystretch{1.1}
		\setlength{\tabcolsep}{1.5mm}{
			\begin{tabular}{c|cc|ccc}
				\hline
				& Orig.       & \#(Param.) & Reshape    & Core     & \#(Param.) \\ \hline \hline
				Conv1 & 32$\times$3$\times$5$\times$5    & 2K                  & 96$\times$25      & 12$\times$12    & 1.6K                \\
				Conv2 & 64$\times$32$\times$5$\times$5   & 51K                 & 64$\times$32$\times$25   & 9$\times$9$\times$9    & 1.9K                \\
				Conv3 & 64$\times$64$\times$5$\times$5   & 102K                & 64$\times$64$\times$25   & 11$\times$11$\times$11 & 3.1K                \\
				Conv4 & 128$\times$64$\times$5$\times$5  & 205K                & 128$\times$64$\times$25  & 11$\times$11$\times$11 & 3.8K                \\
				Conv5 & 256$\times$128$\times$5$\times$5 & 819K                & 256$\times$128$\times$25 & 11$\times$11$\times$11 & 6.1K                \\
				FC1   & 4096$\times$10     & 40K                 & 64$\times$64$\times$10   & 9$\times$9$\times$9    & 2.0K        \\ \hline      
		\end{tabular}}
		\caption{ADA-Tucker setting details on the network architecture used in Sec. \ref{svhn_cifar}.}
		\label{cifar_ada}
	\end{table}
	
	\begin{table}[h]
		\centering
		\small
		\renewcommand\arraystretch{1.1}
		\setlength{\tabcolsep}{1.5mm}{
			\begin{tabular}{cc|cc|ccc}
				\hline
				\multicolumn{2}{c|}{ResNet-20}   & Orig.     & \#(Param.) & Reshape   & Core      & \#(Param.) \\\hline\hline
				\multirow{2}{*}{Block0} & Conv1 & 16$\times$16$\times$3$\times$3 & 2304       & 16$\times$16$\times$9   & 12$\times$12$\times$6   & 1302       \\
				& Conv2 & 16$\times$16$\times$3$\times$3 & 2304       & 16$\times$16$\times$9   & 12$\times$12$\times$6   & 1302       \\
				\multirow{2}{*}{Block1} & Conv1 & 16$\times$16$\times$3$\times$3 & 2304       & 16$\times$16$\times$9   & 12$\times$12$\times$6   & 1302       \\
				& Conv2 & 16$\times$16$\times$3$\times$3 & 2304       & 16$\times$16$\times$9   & 12$\times$12$\times$6   & 1302       \\
				\multirow{2}{*}{Block2} & Conv1 & 16$\times$16$\times$3$\times$3 & 2304       & 16$\times$16$\times$9   & 12$\times$12$\times$6   & 1302       \\
				& Conv2 & 16$\times$16$\times$3$\times$3 & 2304       & 16$\times$16$\times$9   & 12$\times$12$\times$6   & 1302       \\ \hline\hline
				\multirow{2}{*}{Block3} & Conv1 & 32$\times$16$\times$3$\times$3 & 4608       & 18$\times$16$\times$16  & 12$\times$10$\times$10  & 1736       \\
				& Conv2 & 32$\times$32$\times$3$\times$3 & 9216       & 12$\times$12$\times$8$\times$8 & 8$\times$8$\times$6$\times$6   & 2592       \\
				\multirow{2}{*}{Block4} & Conv1 & 32$\times$32$\times$3$\times$3 & 9216       & 12$\times$12$\times$8$\times$8 & 8$\times$8$\times$6$\times$6   & 2592       \\
				& Conv2 & 32$\times$32$\times$3$\times$3 & 9216       & 12$\times$12$\times$8$\times$8 & 8$\times$8$\times$6$\times$6   & 2592       \\
				\multirow{2}{*}{Block5} & Conv1 & 32$\times$32$\times$3$\times$3 & 9216       & 12$\times$12$\times$8$\times$8 & 8$\times$8$\times$6$\times$6   & 2592       \\ 
				& Conv2 & 32$\times$32$\times$3$\times$3 & 9216       & 12$\times$12$\times$8$\times$8 & 8$\times$8$\times$6$\times$6   & 2592       \\ \hline\hline
				\multirow{2}{*}{Block6} & Conv1 & 64$\times$32$\times$3$\times$3 & 18432      & 32$\times$24$\times$24  & 24$\times$16$\times$16  & 7680       \\
				& Conv2 & 64$\times$64$\times$3$\times$3 & 36864      & 9$\times$8$\times$8$\times$8$\times$8 & 6$\times$6$\times$6$\times$6$\times$6 & 8022       \\
				\multirow{2}{*}{Block7} & Conv1 & 64$\times$64$\times$3$\times$3 & 36864      & 9$\times$8$\times$8$\times$8$\times$8 & 6$\times$6$\times$6$\times$6$\times$6 & 8022       \\
				& Conv2 & 64$\times$64$\times$3$\times$3 & 36864      & 9$\times$8$\times$8$\times$8$\times$8 & 6$\times$6$\times$6$\times$6$\times$6 & 8022       \\
				\multirow{2}{*}{Block8} & Conv1 & 64$\times$64$\times$3$\times$3 & 36864      & 9$\times$8$\times$8$\times$8$\times$8 & 6$\times$6$\times$6$\times$6$\times$6 & 8022       \\
				& Conv2 & 64$\times$64$\times$3$\times$3 & 36864      & 9$\times$8$\times$8$\times$8$\times$8 & 6$\times$6$\times$6$\times$6$\times$6 & 8022   
				\\ \hline   
		\end{tabular}}
		
		\caption{ADA-Tucker setting details on ResNet-20 (In Sec. \ref{modern_net}).}
		\label{rnrn20}
	\end{table}

	\begin{table}[]
		\centering
		\small
		\renewcommand\arraystretch{1.3}
		\setlength{\tabcolsep}{0.6mm}{
			\begin{tabular}{cc|cc|ccc}
				\hline
				\multicolumn{2}{c|}{Wide-ResNet-28-10} & Orig.                                                                  & \#(Param.)                & Reshape                         & Core                            & \#(Param.)               \\
				\hline \hline
				\multirow{2}{*}{Block0}                      & Conv1                     & 160$\times$16$\times$3$\times$3                      & 23K                       & 16$\times$16$\times$10$\times$9                      & 10$\times$10$\times$6$\times$6                       & 4K                       \\
				& Conv2                     & 160$\times$160$\times$3$\times$3                     & 230K                      & 24$\times$24$\times$20$\times$20                     & 12$\times$12$\times$12$\times$12                     & 22K                      \\
				\multirow{2}{*}{Block1}                      & Conv1                     & 160$\times$160$\times$3$\times$3                     & 230K                      & 24$\times$24$\times$20$\times$20                     & 12$\times$12$\times$12$\times$12                     & 22K                      \\
				& Conv2                     & 160$\times$160$\times$3$\times$3                     & 230K                      & 24$\times$24$\times$20$\times$20                     & 12$\times$12$\times$12$\times$12                     & 22K                      \\
				\multirow{2}{*}{Block2}                      & Conv1                     & 160$\times$160$\times$3$\times$3                     & 230K                      & 24$\times$24$\times$20$\times$20                     & 12$\times$12$\times$12$\times$12                     & 22K                      \\
				& Conv2                     & 160$\times$160$\times$3$\times$3                     & 230K                      & 24$\times$24$\times$20$\times$20                     & 12$\times$12$\times$12$\times$12                     & 22K                      \\
				\multirow{2}{*}{Block3}                      & Conv1                     & 160$\times$160$\times$3$\times$3                     & 230K                      & 24$\times$24$\times$20$\times$20                     & 12$\times$12$\times$12$\times$12                     & 22K                      \\
				& Conv2                     & 160$\times$160$\times$3$\times$3                     & 230K                      & 24$\times$24$\times$20$\times$20                     & 12$\times$12$\times$12$\times$12                     & 22K                      \\\hline\hline
				\multirow{2}{*}{Block4}                      & Conv1                     & 320$\times$160$\times$3$\times$3                     & 460K                      & 80$\times$80$\times$72                        & 36$\times$36$\times$36                        & 55K                      \\
				& Conv2                     & 320$\times$320$\times$3$\times$3                     & 921K                      & 32$\times$32$\times$30$\times$30                     & 16$\times$16$\times$16$\times$16                     & 67K                      \\
				\multirow{2}{*}{Block5}                      & Conv1                     & 320$\times$320$\times$3$\times$3                     & 921K                      & 32$\times$32$\times$30$\times$30                     & 16$\times$16$\times$16$\times$16                     & 67K                      \\
				& Conv2                     & 320$\times$320$\times$3$\times$3                     & 921K                      & 32$\times$32$\times$30$\times$30                     & 16$\times$16$\times$16$\times$16                     & 67K                      \\
				\multirow{2}{*}{Block6}                      & Conv1                     & 320$\times$320$\times$3$\times$3                     & 921K                      & 32$\times$32$\times$30$\times$30                     & 16$\times$16$\times$16$\times$16                     & 67K                      \\
				& Conv2                     & 320$\times$320$\times$3$\times$3                     & 921K                      & 32$\times$32$\times$30$\times$30                     & 16$\times$16$\times$16$\times$16                     & 67K                      \\
				\multirow{2}{*}{Block7}                      & Conv1                     & 320$\times$320$\times$3$\times$3                     & 921K                      & 32$\times$32$\times$30$\times$30                     & 16$\times$16$\times$16$\times$16                     & 67K                      \\
				& Conv2                     & 320$\times$320$\times$3$\times$3                     & 921K                      & 32$\times$32$\times$30$\times$30                     & 16$\times$16$\times$16$\times$16                     & 67K                      \\ \hline\hline
				\multirow{2}{*}{Block8}                      & Conv1                     & 640$\times$320$\times$3$\times$3                     & 1843K                     & 40$\times$40$\times$36$\times$32                     & 18$\times$18$\times$18$\times$18                     & 108K                     \\
				& Conv2                     & 640$\times$640$\times$3$\times$3                     & 3686K                     & 48$\times$48$\times$40$\times$40                     & 20$\times$20$\times$20$\times$20                     & 163K                     \\
				\multicolumn{1}{c}{\multirow{2}{*}{Block9}}  & \multicolumn{1}{c|}{Conv1} & \multicolumn{1}{c}{640$\times$640$\times$3$\times$3} & \multicolumn{1}{c|}{3686K} & \multicolumn{1}{c}{48$\times$48$\times$40$\times$40} & \multicolumn{1}{c}{20$\times$20$\times$20$\times$20} & \multicolumn{1}{c}{163K} \\
				\multicolumn{1}{c}{}                         & \multicolumn{1}{c|}{Conv2} & \multicolumn{1}{c}{640$\times$640$\times$3$\times$3} & \multicolumn{1}{c|}{3686K} & \multicolumn{1}{c}{48$\times$48$\times$40$\times$40} & \multicolumn{1}{c}{20$\times$20$\times$20$\times$20} & \multicolumn{1}{c}{163K} \\
				\multicolumn{1}{c}{\multirow{2}{*}{Block10}} & \multicolumn{1}{c|}{Conv1} & \multicolumn{1}{c}{640$\times$640$\times$3$\times$3} & \multicolumn{1}{c|}{3686K} & \multicolumn{1}{c}{48$\times$48$\times$40$\times$40} & \multicolumn{1}{c}{20$\times$20$\times$20$\times$20} & \multicolumn{1}{c}{163K} \\
				\multicolumn{1}{c}{}                         & \multicolumn{1}{c|}{Conv2} & \multicolumn{1}{c}{640$\times$640$\times$3$\times$3} & \multicolumn{1}{c|}{3686K} & \multicolumn{1}{c}{48$\times$48$\times$40$\times$40} & \multicolumn{1}{c}{20$\times$20$\times$20$\times$20} & \multicolumn{1}{c}{163K} \\
				\multicolumn{1}{c}{\multirow{2}{*}{Block11}} & \multicolumn{1}{c|}{Conv1} & \multicolumn{1}{c}{640$\times$640$\times$3$\times$3} & \multicolumn{1}{c|}{3686K} & \multicolumn{1}{c}{48$\times$48$\times$40$\times$40} & \multicolumn{1}{c}{20$\times$20$\times$20$\times$20} & \multicolumn{1}{c}{163K} \\
				\multicolumn{1}{c}{}                         & \multicolumn{1}{c|}{Conv2} & \multicolumn{1}{c}{640$\times$640$\times$3$\times$3} & \multicolumn{1}{c|}{3686K} & \multicolumn{1}{c}{48$\times$48$\times$40$\times$40} & \multicolumn{1}{c}{20$\times$20$\times$20$\times$20} & \multicolumn{1}{c}{163K} \\ \hline
		\end{tabular}}
		
		\caption{ADA-Tucker setting details on Wide ResNet-28-10 (In Sec. \ref{modern_net}).}
		\label{wrnwrn}
	\end{table}

\end{appendices}

\end{document}